\documentclass{article}

\PassOptionsToPackage{numbers,sort&compress}{natbib}
\usepackage[preprint]{neurips_2026}

\usepackage[utf8]{inputenc} 
\usepackage[T1]{fontenc}    
\usepackage{hyperref}       
\usepackage{url}            
\usepackage{booktabs}       
\usepackage{amsfonts}       
\usepackage{nicefrac}       
\usepackage{microtype}      
\usepackage{xcolor}         
\usepackage{graphicx}
\usepackage{wrapfig}
\usepackage{subcaption}
\usepackage{array}
\usepackage{multirow}

\usepackage{amsmath}
\usepackage{amssymb}
\usepackage{mathtools}
\usepackage{amsthm}
\usepackage[capitalize,noabbrev]{cleveref}

\usepackage{algorithm}
\usepackage{algorithmic}

\theoremstyle{plain}

\theoremstyle{definition}

\theoremstyle{remark}

\newcommand{\pmstd}[2]{#1\,\text{\scriptsize(#2)}}

\title{naPINN: Noise-Adaptive Physics-Informed \\ Neural Networks for Recovering Physics \\ from Corrupted Measurement}

\author{%
  Hankyeol Kim \qquad Pilsung Kang\thanks{Corresponding author.} \\
  Department of Industrial Engineering\\
  Seoul National University\\
  Seoul, Republic of Korea \\
  \texttt{\{hankyeol,pilsung\_kang\}@snu.ac.kr} \\
}

\begin{document}

\maketitle

\begin{abstract}
    Physics-Informed Neural Networks (PINNs) are effective methods for solving inverse problems and discovering governing equations from observational data. However, their performance degrades under complex measurement noise and gross outliers. To address this issue, we propose the Noise-Adaptive Physics-Informed Neural Network (\textbf{naPINN}), which robustly recovers physical solutions from corrupted measurements without prior knowledge of the noise distribution. naPINN estimates residual reliability during training and uses a trainable reliability gate to downweight unreliable measurement data, while a rejection-cost regularizer prevents trivial solutions where valid data are discarded. We demonstrate the efficacy of naPINN on benchmark partial differential equations corrupted by non-Gaussian noise and varying rates of outliers. The results show that naPINN improves robustness over existing robust PINN baselines, while isolating outliers and accurately reconstructing the dynamics under severe data corruption.
\end{abstract}

\section{Introduction}
\label{sec:introduction}

Modeling physical systems governed by partial differential equations (PDEs) has traditionally relied on classical numerical solvers such as finite element, finite difference, and finite volume methods \cite{FEM-book, FDM-book}. While highly effective for forward simulations under fully specified conditions, these mesh-based approaches face fundamental challenges in inverse problems, where latent states or system parameters must be inferred from sparse, indirect, and noisy measurements. Such problems are often ill-posed and highly sensitive to data corruption, limiting the robustness of conventional solvers \cite{CptlMethods-Book, StatCom-Book}.

Physics-Informed Neural Networks (PINNs) have emerged as a powerful alternative within scientific machine learning by embedding PDE constraints directly into neural network training via automatic differentiation \cite{Raissi2019, Baydin2018}. Owing to their mesh-free formulation and ability to incorporate scattered observations, PINNs have shown promise for inverse problems, parameter identification, and data assimilation across a wide range of scientific domains \cite{Karniadakis2021}.

Despite their promise, standard PINN formulations are sensitive to corrupted measurement data. This is a central obstacle for real-world inverse problems: measurements are often sparse, boundary or initial conditions may be unavailable, and the data can be contaminated by sensor faults, transmission errors, calibration drift, or environmental interference \cite{Chander2022}. Nevertheless, much of the PINN literature evaluates inverse problems under clean observations or weak, well-behaved Gaussian perturbations \cite{Pilar2024}. Such assumptions are convenient because the standard mean squared error (MSE) data loss corresponds to a Gaussian noise model, but they are mismatched to practical sensing regimes in which noise may be asymmetric, multimodal, heavy-tailed, or mixed with gross outliers. In these settings, a small fraction of corrupted measurements can dominate gradients and bias the recovered solution or PDE parameters.

Existing robust and uncertainty-aware PINN approaches address parts of this problem, but they do not directly infer measurement reliability under unknown corruption. Fixed robust penalties or likelihoods reduce sensitivity to large residuals, while Bayesian PINNs primarily quantify posterior uncertainty rather than perform measurement-level selection \cite{Peng2022, Duarte2025, Yang2021}. Outside scientific machine learning, selective learning and test-time adaptation suggest that learning can be more stable when updates emphasize reliable samples \cite{Yang2024,Wang2024DSS,Kim2024}. This principle has not been developed for measurement-driven inverse PINNs, where sample reliability must be inferred jointly with a PDE-constrained solution and unknown physical parameters.

In this work, we address this gap by proposing \textbf{naPINN} (Noise-Adaptive Physics-Informed Neural Networks), a framework for measurement-driven inverse PDE problems with corrupted observations and no assumed parametric noise law. The key idea is to estimate the reliability of measurement residuals during training and use this signal to adaptively regulate the influence of individual data points. After a warm-up stage, residuals from a reasonably aligned predictor become informative about measurement corruption: residual values that are unlikely under the learned noise distribution are treated as less reliable and are more likely to correspond to noisy or outlying measurements. naPINN is built around a modular residual-based noise distribution estimator, which can be instantiated by nonparametric, mixture-based, or energy-based density estimators. In our main instantiation, we use a one-dimensional Energy-Based Model (EBM) \cite{LeCun2006} as a representative flexible estimator. The resulting residual reliability score is converted into a trainable gate that downweights measurements with anomalous behavior, while a rejection-cost regularizer prevents degenerate solutions in which the model discards valid data. To our knowledge, naPINN is the first inverse-PINN framework to address corrupted measurements with unknown noise distributions by coupling residual-based noise distribution estimation with trainable per-measurement reliability gating.

Our contributions can be summarized as follows:

\begin{itemize}
    \item We identify measurement-driven inverse PINNs under unknown measurement-noise distributions and gross outliers as an important but still insufficiently addressed setting for robust physics recovery.
    \item We propose \textbf{naPINN}, a modular reliability-gated PINN framework that couples residual-based noise distribution estimation with adaptive measurement selection through a reliability gate with rejection cost regularization. The main experiments use an EBM estimator, while alternative density-estimator components are evaluated in the appendix.
    \item We validate the proposed framework on 2D PDE benchmarks under multimodal non-Gaussian noise and varying outlier ratios. The results demonstrate that naPINN improves robustness over the evaluated PINN and robust PINN baselines, accurately reconstructing dynamics while explicitly identifying and suppressing corrupted measurements.
\end{itemize}

\section{Related work}
\label{sec:relatedwork}

\paragraph{Stabilizing PINN optimization.}
A large line of work addresses optimization pathologies in PINNs (e.g., stiffness, spectral bias, and gradient imbalance) via adaptive loss reweighting, gradient-based balancing, or curriculum-like scheduling \cite{Rahaman2019, Wang2021, SWang2022, CWang2022, Chen2018, Xiang2022, Hwang2024, Bischof2025}.  
Related techniques also include adaptive sampling \cite{Nabian2021} and region-based optimization \cite{Wu2024, duan2025} to stabilize PINN training. Some works assign adaptive weights to data, boundary, or residual terms \cite{McClenny2023}, but these weights are usually designed to balance optimization terms or sampling regions rather than to infer whether each noisy measurement data is reliable.

\paragraph{Robust learning from corrupted measurements in PINNs.}
Robust PINN variants usually modify the data loss. LAD-PINN and MAD-PINN use $\ell_1$ fitting and median-based screening to reduce the influence of corrupted observations \cite{Peng2022}. OrPINN uses a $q$-Gaussian likelihood as an outlier-resistant loss \cite{Duarte2025}. These methods are useful baselines, but they rely on a fixed penalty or a chosen likelihood family. They do not estimate a noise distribution from residuals and then use residual reliability to decide the contribution of each measurement during training. B-PINNs infer posterior uncertainty in forward and inverse PDE problems \cite{Yang2021}, and related adversarial or probabilistic formulations also focus on uncertainty quantification \cite{Yang2019}. Uncertainty estimates are not the same as measurement selection: a high-uncertainty observation is not necessarily removed or downweighted, and a corrupted observation can still affect the likelihood.

\paragraph{Anomaly detection, selective learning, and test-time adaptation beyond SciML.}
Learning under corrupted supervision often relies on sample selection or reweighting using confidence, disagreement, or memorization dynamics \cite{Jiang2018, Han2018, Li2020, Natarajan2013}. Anomaly detection provides another source of reliability scores, commonly based on one-class objectives, reconstruction errors, density estimates, or other rarity measures \cite{Chandola2013, Ruff2018, Pang2021}. In test-time adaptation, robustness studies show that indiscriminate updates on corrupted or shifted samples can be harmful, motivating confidence- or calibration-based filtering before adaptation \cite{Hendrycks2019, Ovadia2019, Mao2022, Yang2024, Wang2024DSS}. Time-series work uses related ideas for anomaly detection, nonstationary forecasting, and adaptation by identifying normal or reliable segments before updating a model \cite{Su2019, Audibert2020, Xu2021, Tuli2022, Kim2024, Fu2025, Wu2025}. These methods motivate reliability-aware learning, but they are not designed for inverse PDE problems where the model must satisfy physics constraints and recover unknown physical parameters.

\begin{figure*}
  \centering
    \includegraphics[width=\textwidth]{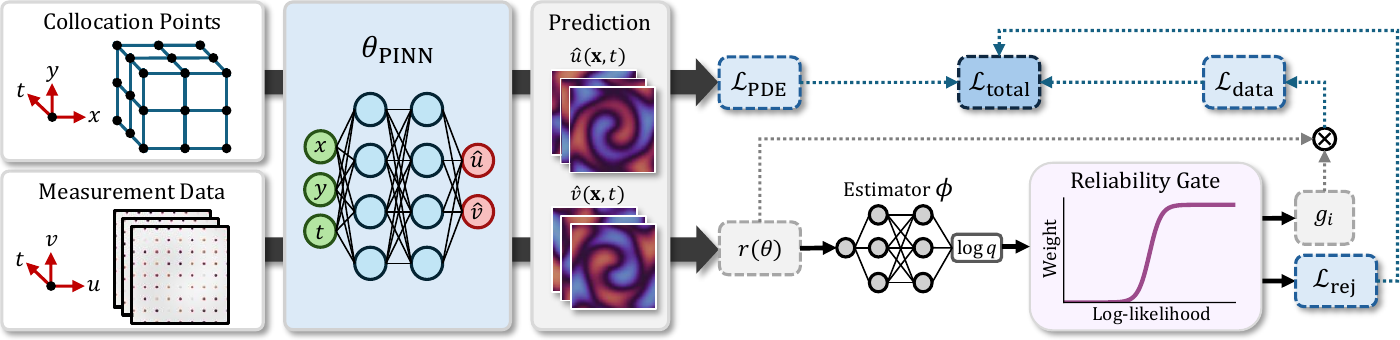}
    \caption{Overview of the naPINN framework. A residual-based noise distribution estimator, which is an EBM in the main instantiation, assigns reliability scores to measurement residuals under the learned noise distribution. A trainable reliability gate converts these scores into adaptive weights, selectively filtering unreliable measurement data when computing the data loss.}
    \label{fig:naPINN}
    \vspace{-0.7\baselineskip}
\end{figure*}

\section{Method}
\label{sec:method}

We propose \textbf{naPINN} (Noise-Adaptive Physics-Informed Neural Networks), a training framework for inverse problems with corrupted measurement data. As summarized in Algorithm~\ref{alg:napinn}, naPINN augments standard PINN training with a residual-based noise distribution estimator and a trainable reliability gate. The estimator converts measurement residuals into residual reliability scores, and the gate uses these scores to determine how strongly each measurement should influence the inverse problem. The overall structure of the framework is illustrated in Figure~\ref{fig:naPINN}.

\subsection{PINNs for inverse problems}
\label{subsec:problem_setup}

We consider a general nonlinear partial differential equation (PDE) parameterized by $\boldsymbol{\lambda}$, defined over a spatiotemporal domain $\Omega \subset \mathbb{R}^d$ and a time interval $t \in [0, T]$:
\begin{equation}
    \mathcal{N}[u](\mathbf{x},t;\boldsymbol{\lambda}) = f(\mathbf{x},t), \quad (\mathbf{x},t) \in \Omega \times [0, T] ,
\end{equation}
where $u(\mathbf{x},t)$ denotes the latent solution, $\mathcal{N}[\cdot]$ is a general differential operator, and $f(\mathbf{x},t)$ is a forcing term. The system is subject to boundary conditions (BC) on $\partial \Omega$ and initial conditions (IC) at $t=0$.
Standard PINN frameworks aim to approximate the solution $u(\mathbf{x},t)$ using a deep neural network $\hat{u}_{\theta}(\mathbf{x},t)$ with trainable parameters $\theta$. The network is trained by minimizing a composite loss function that enforces both the physics and the available observations:
\begin{equation}
\begin{aligned}
\mathcal{L}_\mathrm{PINN}(\theta, \boldsymbol{\lambda})
&= w_f \mathcal{L}_\mathrm{PDE} + w_b \mathcal{L}_\mathrm{BC} + w_i \mathcal{L}_\mathrm{IC} + w_d \mathcal{L}_\mathrm{data}.
\end{aligned}
\label{eq:standard_pinn}
\end{equation}
where $w_f$, $w_b$, $w_i$, and $w_d$ are weights for the corresponding terms. Here, $\mathcal{L}_\mathrm{PDE}$ penalizes residuals of the governing equation on a set of collocation points $\{ ( \mathbf{x}_f^{(i)}, t_f^{(i)} ) \}_{i=1}^{N_f}$, computed via automatic differentiation:
\begin{equation}
    \mathcal{L}_\mathrm{PDE} = \frac{1}{N_f} \sum _{i=1} ^{N_f} \left\| \mathcal{N}[\hat{u}_\theta] (\mathbf{x}_f^{(i)}, t_f^{(i)}; \boldsymbol{\lambda}) - f(\mathbf{x}_f^{(i)}, t_f^{(i)}) \right\|^2 .
\label{eq:pde_loss}
\end{equation}
We focus on a measurement-driven inverse setting in which boundary and initial conditions are unavailable or incomplete, and the model must infer the solution and unknown PDE parameters from noisy measurements $\mathcal{D}_d=\{(\mathbf{x}_d^{(i)},t_d^{(i)},y_d^{(i)})\}_{i=1}^{N_d}$. Each observation satisfies $y_d^{(i)}=y_t^{(i)}+\epsilon_i$, where $y_t^{(i)}$ is the clean solution value and $\epsilon_i$ is unknown measurement noise. The standard data term is the mean squared error
\begin{equation}
\mathcal{L}_\mathrm{data}
= \frac{1}{N_d} \sum_{i=1}^{N_d}
\left\| \hat{u}_\theta(\mathbf{x}_d^{(i)}, t_d^{(i)}) - y_d^{(i)} \right\|^2 .
\label{eq:data_loss}
\end{equation}
This loss corresponds to maximum likelihood under Gaussian measurement noise; when the true corruption is multimodal, asymmetric, heavy-tailed, or mixed with gross outliers, the resulting gradients can be dominated by unreliable measurement data. For each measurement, we define the data residual as
\begin{equation}
r^{(i)}(\theta) := y_d^{(i)} - \hat{u}_\theta(\mathbf{x}_d^{(i)},t_d^{(i)}).
\label{eq:residual_def}
\end{equation}

\begin{algorithm}[t]
\caption{Noise-Adaptive PINN (naPINN)}
\label{alg:napinn}
\begin{algorithmic}[1]
\STATE \textbf{Input:} total iters $K_{\text{tot}}$, warm-up iters $K_{w}$,
estimator init iters $K_{\text{est}}$, datasets $\mathcal{D}_c, \mathcal{D}_d$,
operator $\mathcal{N}(\boldsymbol{\lambda})$, rejection-cost weight $\lambda_{\text{rej}}$,
EMA decay $\rho$
\STATE \textbf{Output:} optimized $\theta_\text{PINN}$, PDE parameters $\boldsymbol{\lambda}$, estimator state $\phi$, gate parameters $(a, \tau)$
\STATE \textbf{Initialize:} $\theta_\text{PINN}, \boldsymbol{\lambda}, \phi, (a,\tau)$, running std $\sigma_{\text{run}}$
\STATE \textit{// Phase 1: PINN warm-up}
\FOR{$i = 1, \dots, K_{w}$}
    \STATE sample minibatches from $\mathcal{D}_c, \mathcal{D}_d$
    \STATE update $(\theta_\text{PINN}, \boldsymbol{\lambda})$ by minimizing $\mathcal{L}_{\text{PINN}}$ \hfill \COMMENT{Eq.~\eqref{eq:standard_pinn}}
\ENDFOR
\STATE \textit{// Phase 2: estimator initialization (skip if $\phi$ has no trainable params)}
\STATE compute residuals $\{r_i\}$ on $\mathcal{D}_d$ with current $\theta_\text{PINN}$
\FOR{$k = 1, \dots, K_{\text{est}}$}
    \STATE sample residual minibatch; update $\sigma_{\text{run}}$ via EMA \hfill \COMMENT{Eq.~\eqref{eq:ema_std}}
    \STATE normalize $\tilde r_i \leftarrow r_i/\sigma_{\text{run}}$; update $\phi$ on $\{\tilde r_i\}$
\ENDFOR
\STATE \textit{// Phase 3: joint training with reliability gating}
\FOR{$i = K_{w}+1, \dots, K_{\text{tot}}$}
    \STATE sample minibatches; compute residuals and update $\sigma_{\text{run}}$
    \STATE compute scores $s_\phi(\tilde r_i)$ and gate weights $g_i$ \hfill \COMMENT{Eq.~\eqref{eq:gate}}
    \STATE update $(\theta_\text{PINN}, \boldsymbol{\lambda}, a, \tau)$ by minimizing $\mathcal{L}_{\text{total}}$ \hfill \COMMENT{Eq.~\eqref{eq:joint_pinn_gate}}
    \STATE update $\phi$ on $\{\tilde r_i\}$
\ENDFOR
\end{algorithmic}
\end{algorithm}

\subsection{Residual reliability estimation}
\label{subsec:residual_reliability}
naPINN estimates residual reliability from normalized residuals $\tilde r_i$. During early training, residuals may reflect both prediction error and measurement noise. After warm-up, however, residuals become a more useful proxy for identifying corrupted observations: residuals falling in low-probability regions of the learned noise distribution are treated as less reliable. We represent the residual-based noise distribution estimator by a state or parameter vector $\phi$ and define its scalar reliability score as $s_\phi(\tilde r_i)$, where larger values indicate lower residual reliability. In general, if the estimator provides a density $\hat p_\phi$, we use the negative log-density, $s_\phi(\tilde r_i)=-\log \hat p_\phi(\tilde r_i)$. This interface covers nonparametric density fitting, mixture-based estimators, and energy-based scoring without changing the downstream gate, including Kernel Density Estimation (KDE), Gaussian Mixture Model (GMM), and Energy-Based Model (EBM) variants. In the main experiments, we use a one-dimensional Energy-Based Model (EBM) as the representative estimator. It defines an unnormalized density
\begin{equation}
p_\phi(r) = \frac{\exp(-E_\phi(r))}{Z_\phi}, \qquad
Z_\phi = \int \exp(-E_\phi(r))\,dr.
\label{eq:ebm_density}
\end{equation}
For this instantiation, the reliability score is the energy, $s_\phi(\tilde r_i)=E_\phi(\tilde r_i)$, which is equivalent to the negative log-density up to an additive constant.

\subsection{Staged warm-up and estimator initialization}
\label{subsec:stage1}
Since early residuals mix prediction error with measurement noise, naPINN uses a short staged procedure before full joint training. We first train the PINN parameters, and the PDE parameters $\boldsymbol{\lambda}$ when applicable, with the standard objective in Eq.~\eqref{eq:standard_pinn}:
\begin{equation}
(\theta,\boldsymbol{\lambda}) \leftarrow 
\arg\min_{\theta,\boldsymbol{\lambda}} \ \mathcal{L}_{\text{PINN}}(\theta,\boldsymbol{\lambda}),
\label{eq:warmup}
\end{equation}
which yields a predictor $u_{\theta^{(0)}}$ whose residuals better reflect the corruption structure in the measurements. We then freeze the warmed-up PINN and compute measurement residuals:
\begin{equation}
r_i = y_i - u_{\theta^{(0)}}(\mathbf{x}_i,t_i), \qquad i=1,\dots,N_r,
\end{equation}
where $N_r$ denotes the number of sampled residuals. We normalize these residual values and initialize the noise distribution estimator $\phi$ through the estimator-specific interface used throughout training. A nonparametric estimator is fitted or refreshed on residual samples, whereas a trainable density model is optimized by its likelihood objective. For the EBM used in the main experiments, this corresponds to approximate maximum-likelihood training of $p_\phi(r)\propto \exp(-E_\phi(r))$ on normalized residuals. This initialization stage provides a more stable starting point of residual distribution for the estimator; its effect is evaluated in Appendix~\ref{appendix:ablation}.

Residual magnitudes can vary substantially across minibatches under heavy-tailed noise or gross outliers. Since the reliability estimator operates on normalized residuals, using only the current minibatch scale can make the scores sensitive to rare extreme samples. We therefore maintain a running standard deviation $\sigma_{\text{run}}$ using an exponential moving average (EMA). For a residual minibatch $\mathcal{B}$ with empirical standard deviation $s_{\mathcal{B}}$, we update
\begin{equation}
\sigma_{\text{run}} \leftarrow (1-\beta)\,\sigma_{\text{run}} + \beta\, s_{\mathcal{B}},
\label{eq:ema_std}
\end{equation}
and normalize residuals as $\tilde r = r / \sigma_\text{run}$, where $\beta\in(0,1)$ is a constant. This running-statistics normalization reduces sensitivity to outlier-dominated minibatches and improves convergence during estimator initialization and subsequent joint optimization; the ablation is reported in Appendix~\ref{appendix:running_std_ablation}.

\subsection{Joint optimization with reliability gating}
\label{subsec:stage2}
We then jointly optimize the PINN, the noise distribution estimator, and the reliability gate. The estimator continues to be updated or refreshed on normalized residuals through the same interface described above. As the physical predictor improves, residual reliability becomes a sharper proxy for measurement corruption; in turn, the gate suppresses measurements whose residuals are unlikely under the learned noise distribution.

\paragraph{Reliability gate.}
For each $\tilde r_i$, the estimator outputs a score $s_\phi(\tilde r_i)$. To make the cutoff scale-invariant across estimators and training stages, we standardize scores within each minibatch, $z_i := (s_\phi(\tilde r_i) - \mu_s)/\sigma_s$, and define
\begin{equation}
g_i := \sigma\!\big(a(\tau - z_i)\big),
\label{eq:gate}
\end{equation}
where $a > 0$ is steepness and $\tau$ is cutoff, both trainable (with $a$ reparameterized via softplus). Larger scores yield smaller weights, so measurements falling in low-density regions of $p_\phi$ are downweighted.
\begin{equation}
\mathcal{L}_{d}^{\text{gate}}(\theta)
= \frac{1}{N_d}\sum_{i=1}^{N_d} g_i \, \|y_i - u_\theta(\mathbf{x}_i,t_i)\|_2^2.
\label{eq:gated_data}
\end{equation}

\paragraph{Rejection-cost regularization.}
If unconstrained, the gate may converge to a trivial solution that rejects most measurements to reduce $\mathcal{L}_{d}^{\text{gate}}$. To discourage excessive rejection, we introduce a rejection cost regularization and weight it by $\lambda_{\text{rej}}>0$:
\begin{equation}
\mathcal{L}_{\text{rej}}
= \frac{1}{N_d}\sum_{i=1}^{N_d} (1-g_i).
\label{eq:rej_cost}
\end{equation}

\paragraph{Training objective.}
In this phase, we optimize the PINN and gate parameters by minimizing
\begin{equation}
\mathcal{L}_{\text{total}}
= \mathcal{L}_{\text{PDE}}
+ \mathcal{L}_{d}^{\text{gate}}
+ \lambda_{\text{rej}}\mathcal{L}_{\text{rej}}.
\label{eq:joint_pinn_gate}
\end{equation}
The residual estimator is updated or refreshed on current normalized residuals using the estimator interface described above, with estimator-specific implementation details reported in Appendix~\ref{appendix:ImplementationDetails}.

\paragraph{Connection to a latent-variable robust likelihood.}
The gated objective admits a probabilistic interpretation. Suppose measurement residuals follow a two-component mixture, $r_i \sim \pi, p_\phi(r) + (1-\pi), q(r)$, where $p_\phi$ is the clean-residual density learned by the estimator and $q$ is an unknown outlier density. Under this model, the posterior probability that observation $i$ is clean is
$$\gamma_i ;=; \frac{\pi, p_\phi(r_i)}{\pi, p_\phi(r_i) + (1-\pi), q(r_i)},$$
and the corresponding posterior-weighted log-likelihood takes the form $\sum_i \gamma_i \log p_\phi(r_i)$, which is structurally identical to the gate-weighted data loss in Eq.~\eqref{eq:gated_data}. The naPINN gate $g_i$ plays the role of $\gamma_i$ but is computed from $p_\phi$ and a learned cutoff $\tau$ rather than from a hand-specified $q$, while the rejection cost $\lambda_{\text{rej}}\sum_i (1-g_i)$ acts as a Beta-type prior favoring high inclusion rates. This view explains why naPINN does not require parametric assumptions on $q$: $\tau$ adapts to whichever density level separates clean residuals from anomalies under $p_\phi$.

\begin{sc}
\begin{table*}[t]
\centering
\caption{Performance comparison under corrupted measurements across three outlier ratios. Results are averaged over 10 independent trials with different random seeds, with standard deviations shown in parentheses. The best results are highlighted in bold and the second best results are underlined. \textit{Improvement} refers to the relative improvement of naPINN over the second-best performing baseline. \textit{Clean measurement} row reports rMAE and rMSE errors when training the PINN on clean measurement data without noise or outliers.}
\label{table:main_result}
\setlength{\tabcolsep}{4pt} 
\resizebox{\textwidth}{!}{%
\begin{tabular}{llccccccccc}
\toprule
 & & \multicolumn{3}{c}{Allen--Cahn} & \multicolumn{3}{c}{Burgers} & \multicolumn{3}{c}{$\lambda$--$\omega$ RD} \\
\cmidrule(lr){3-5}\cmidrule(lr){6-8}\cmidrule(lr){9-11}
Method & Metric & 5\% & 10\% & 15\% & 5\% & 10\% & 15\% & 5\% & 10\% & 15\% \\
\midrule

\multirow{2}{*}{PINN}
 & rMAE & 0.305 & 0.584 & 0.823 & 0.221 & 0.387 & 0.568 & 0.161 & 0.273 & 0.437 \\
 & rMSE & 0.267 & 0.488 & 0.682 & 0.222 & 0.377 & 0.547 & 0.178 & 0.292 & 0.457 \\
\cmidrule(lr){2-11}

\multirow{2}{*}{B-PINN}
 & rMAE & 0.367 & 0.608 & 0.885 & 0.278 & 0.429 & 0.592 & 0.190 & 0.314 & 0.490 \\
 & rMSE & 0.338 & 0.538 & 0.762 & 0.285 & 0.419 & 0.571 & 0.222 & 0.339 & 0.517 \\
\cmidrule(lr){2-11}

\multirow{2}{*}{LAD-PINN}
 & rMAE & 0.272 & 0.318 & 0.352 & 0.203 & 0.225 & 0.242 & 0.155 & 0.172 & 0.190 \\
 & rMSE & 0.233 & 0.272 & 0.296 & 0.194 & 0.216 & 0.241 & 0.164 & 0.181 & 0.199 \\
\cmidrule(lr){2-11}

\multirow{2}{*}{OrPINN ($q=1.9$)}
 & rMAE & 0.249 & 0.450 & 0.654 & 0.153 & 0.258 & 0.376 & \underline{0.093} & \underline{0.130} & 0.178 \\
 & rMSE & 0.219 & 0.379 & 0.545 & 0.158 & 0.267 & 0.373 & \underline{0.109} & \underline{0.150} & 0.201 \\
\cmidrule(lr){2-11}

\multirow{2}{*}{OrPINN ($q=2.9$)}
 & rMAE & \underline{0.165} & \underline{0.219} & \underline{0.269} & \underline{0.130} & \underline{0.155} & \underline{0.175} & 0.123 & 0.138 & \underline{0.150} \\
 & rMSE & \underline{0.151} & \underline{0.195} & \underline{0.234} & \underline{0.132} & \underline{0.155} & \underline{0.180} & 0.140 & 0.151 & \underline{0.163} \\
\midrule

\multirow{2}{*}{\textbf{naPINN (ours)}}
 & rMAE & \textbf{0.104} & \textbf{0.110} & \textbf{0.134} & \textbf{0.075} & \textbf{0.074} & \textbf{0.072} & \textbf{0.073} & \textbf{0.074} & \textbf{0.076} \\
 & rMSE & \textbf{0.101} & \textbf{0.108} & \textbf{0.127} & \textbf{0.089} & \textbf{0.092} & \textbf{0.091} & \textbf{0.092} & \textbf{0.092} & \textbf{0.095} \\
\cmidrule(lr){2-11}

\multirow{2}{*}{\textit{Improvement}}
 & rMAE & 37.97\% & 49.77\% & 50.19\% & 42.31\% & 50.32\% & 58.86\% & 21.51\% & 43.08\% & 49.33\% \\
 & rMSE & 33.11\% & 44.62\% & 45.73\% & 32.58\% & 40.65\% & 49.44\% & 15.60\% & 38.67\% & 41.72\% \\
\midrule

\multirow{2}{*}{\textit{Clean measurement}}
 & rMAE & \multicolumn{3}{c}{0.066} & \multicolumn{3}{c}{0.028} & \multicolumn{3}{c}{0.032} \\
 & rMSE & \multicolumn{3}{c}{0.069} & \multicolumn{3}{c}{0.041} & \multicolumn{3}{c}{0.047} \\
\bottomrule
\end{tabular}%
}
\end{table*}
\end{sc}

\section{Experiments}
\label{sec:experiments}

We evaluate the effectiveness of \textbf{naPINN} on three canonical time-dependent two-dimensional PDE benchmarks: the 2D Burgers' equation, the 2D Allen--Cahn equation, and the 2D $\lambda$--$\omega$ reaction--diffusion (RD) system. These problems are used to construct a controlled sparse-measurement setting motivated by sensor deployments, where noisy observations are collected over a spatiotemporal domain $\Omega \times [0, T] \subset \mathbb{R}^3$. This setting requires the model to recover evolving spatial structures while learning from sparsely observed, corrupted measurements. Detailed formulations and numerical solvers used to generate reference solutions are provided in Appendix~\ref{appendix:DatasetDetails}.

\subsection{Experimental setup}

Following a twin-experiment-style protocol widely used in data assimilation and inverse-problem evaluation, we generate observations from known reference solutions so that reconstruction error, parameter error, and corruption severity can be measured under controlled conditions \cite{Asch2016}. This protocol is also consistent with related noisy-data and robust-PINN evaluations, which assess inverse or physics-informed learning using synthetic noisy or corrupted observations \cite{Yang2021,Peng2022,Duarte2025}. To our knowledge, no widely adopted public benchmark simultaneously provides sparse corrupted sensor measurements, governing PDEs, clean reference fields, unknown parameters, and corruption labels, which are the joint ingredients needed for a controlled evaluation of measurement-driven inverse PINNs under unknown corruption. We therefore interpret our experiments as controlled stress tests under intentionally challenging multimodal non-Gaussian noise and gross outliers, rather than as a substitute for deployment-scale real-world validation.

\paragraph{Data generation.} We construct synthetic datasets contaminated by non-Gaussian noise and gross outliers. For the 2D Allen--Cahn equation, an analytical solution is available and used as ground truth. For the 2D Burgers' and $\lambda$--$\omega$ RD systems, high-fidelity numerical simulations are employed. As detailed in Appendix~\ref{appendix:DatasetDetails}, the Burgers' and $\lambda$--$\omega$ RD systems require predicting two state variables $(u, v)$, whereas the Allen--Cahn equation involves a single scalar field. To reflect practical sensing constraints, we assume sensors are deployed on a fixed $15 \times 15$ spatial grid, each recording measurements at all time steps. In contrast, PDE collocation points are sampled from a denser $100 \times 100$ grid. This design creates a controlled data-scarce regime, as acquiring high-resolution measurement data is often costly or infeasible in practice. The spatiotemporal domain and simulation horizon are chosen separately for each benchmark to adequately capture the characteristic dynamics of the system.

\begin{figure*}[t]
  \centering
    \includegraphics[width=\textwidth]{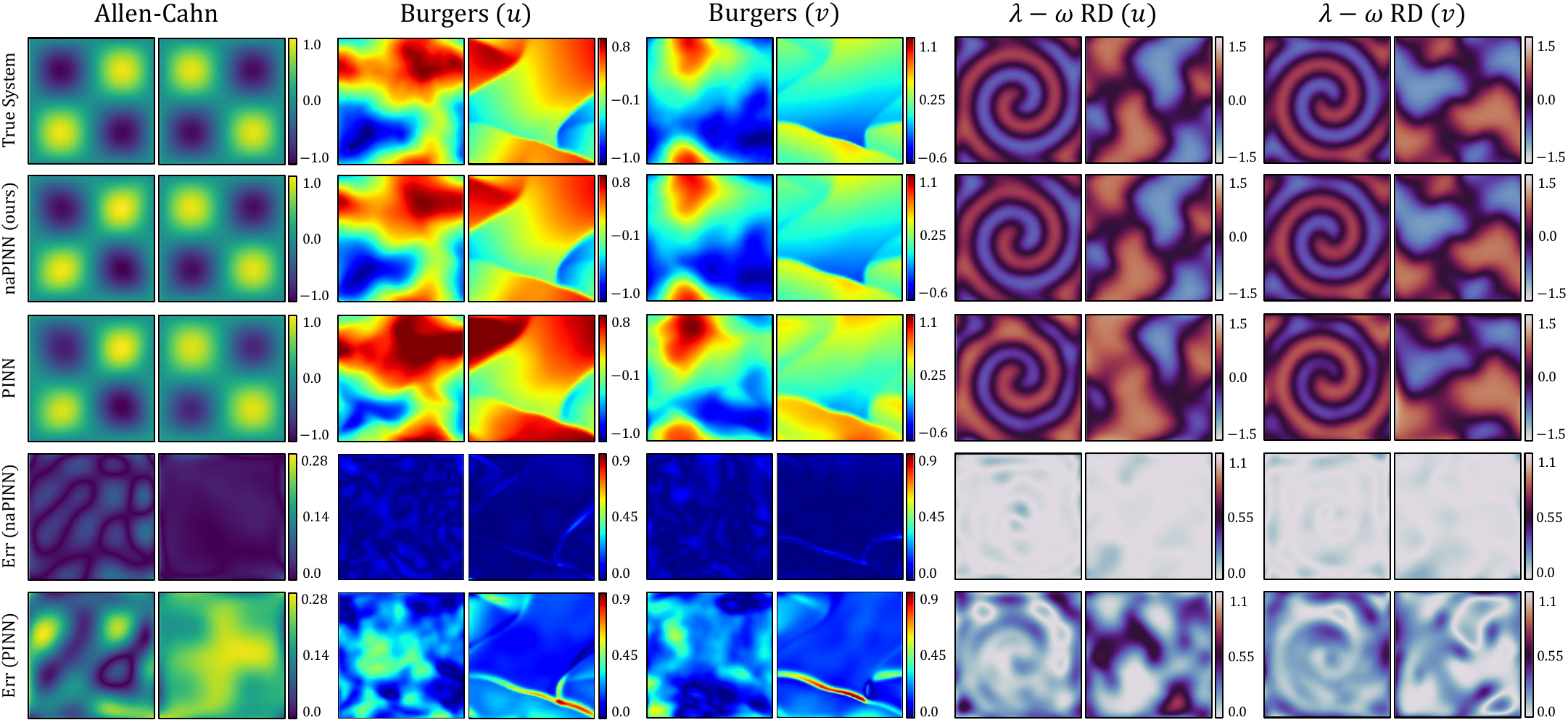}
    \caption{Qualitative comparison of solution reconstruction under corrupted measurements. For each PDE benchmark, we visualize the ground-truth solution, predictions from naPINN and a standard PINN, and their corresponding absolute errors at two representative time steps. Different colormaps are used across benchmarks to enhance visibility.}
    \label{fig:visualization}
    \vspace{-1.0\baselineskip}
\end{figure*}

The same noise distribution is applied across all benchmarks for consistency. The GMM components are specified by $(\mu, \sigma) \in \{(-9.0, 2.0), (-0.3, 4.0), (2.7, 0.6), (8.5, 1.0)\}$, and the overall noise scale is normalized to $10\%$ of the mean absolute magnitude of the corresponding solution field. Gross outliers are introduced by randomly selecting a subset of spatiotemporal measurement points and replacing their values with samples drawn uniformly from a disjoint high-magnitude interval $[k_1 \sigma, k_2 \sigma]$, where $k_1=3$ and $k_2=10$. Compared to prior work~\cite{Duarte2025}, this setting is intentionally more challenging, as the injected outliers deviate less drastically from the base noise distribution. We evaluate robustness under three outlier ratios: $5\%$, $10\%$, and $15\%$ of the total measurement data. Unknown PDE parameters are initialized far from their ground-truth values for all methods to prevent near-true warm-starts; specific initializations are summarized in Appendix~\ref{appendix:DatasetDetails}.

\paragraph{Baselines and metrics.} We compare naPINN against a standard Vanilla PINN and existing robust frameworks. We additionally include B-PINN, implemented with Bayesian linear layers and a mean-field Gaussian weight prior ($\mu=0, \sigma=1$). In particular, for OrPINN \cite{Duarte2025}, which employs a q-Gaussian likelihood, we report results for $q \in \{1.9, 2.9\}$. Reconstruction performance is quantified using relative mean absolute error (rMAE) and relative root mean squared error (rMSE) computed against the clean reference solution on a held-out test set. To assess generalization as a mesh-free PDE solver, all evaluation metrics are computed on a dense $120 \times 120$ spatial grid, which is finer than the collocation grid used during training. Formal definitions of all evaluation metrics are provided in Appendix~\ref{appendix:ImplementationDetails}. For reference, we additionally report results of a standard PINN trained on clean (noise-free) measurements. To isolate the contribution of naPINN, we do not employ auxiliary loss-balancing schemes that dynamically reweight data and PDE residual losses.

\begin{figure*}[t]
  \centering
  \includegraphics[width=\textwidth]{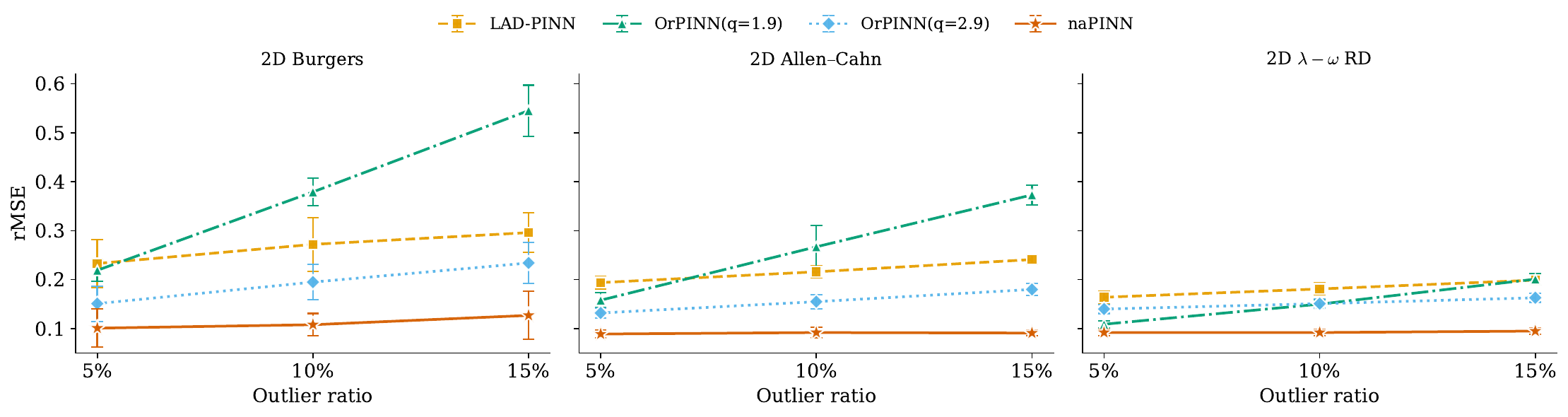}
  \caption{Robustness comparison under increasing outlier ratios on three PDE benchmarks. rMSE is plotted with error bars denoting one standard deviation over independent runs.}
  \label{fig:perf_over_ratios}
  \vspace{-0.9\baselineskip}
\end{figure*}

\subsection{Main results}

Table~\ref{table:main_result} summarizes reconstruction accuracy across all benchmarks and outlier ratios. All methods share a fixed training budget of 30{,}000 steps; per-method schedules and other hyperparameters are detailed in Appendix~\ref{appendix:training_hp}. Results are averaged over 10 independent trials per setting.

Under corrupted measurements, standard PINNs exhibit large rMAE and rMSE values, as squared losses are dominated by gross outliers. Performance degrades rapidly as the outlier ratio increases, often leading to unstable or catastrophic failures. Robust losses such as $L_1$ and $q$-Gaussian significantly improve stability, with the $q$-Gaussian loss at $q=2.9$ yielding the strongest baseline performance in most cases. In contrast, the Bayesian PINN baseline performs poorly across benchmarks, likely because its practical formulation assumes a Gaussian likelihood for tractable inference, which becomes severely misspecified under heavy-tailed, non-Gaussian corruption and gross outliers.

Across all benchmarks and noise levels, naPINN consistently achieves the best performance, with error rates remaining close to those obtained under clean measurements. At the most severe 15\% outlier setting, naPINN substantially reduces rMSE relative to the strongest baseline on all benchmarks. Notably, naPINN exhibits only marginal degradation as the outlier ratio increases from $5\%$ to $15\%$, highlighting the effectiveness of its adaptive reliability gating even in spatiotemporal problems. Beyond field reconstruction, naPINN also recovers the unknown PDE coefficient more accurately than the vanilla baseline; we report parameter-reconstruction results on the Allen--Cahn benchmark in Appendix~\ref{appendix:parameter_reconstruction}.

Although the main naPINN experiments adopt the EBM estimator and the MSE data loss, the reliability-gated framework is not restricted to this choice. Robust losses used by existing baselines can also be combined with naPINN by replacing the data-loss term while keeping the residual estimator and reliability gate unchanged. As shown in Appendix~\ref{appendix:additional_noise_distributions}, for certain corruption distributions, adopting robust losses such as $L_1$ or $q$-Gaussian within the naPINN framework can provide additional performance gains. Appendix~\ref{appendix:estimator_comparison} reports module-swap results with KDE and GMM components, illustrating that naPINN can accommodate multiple residual-density components within the same pipeline. Beyond the data loss and residual estimator, naPINN, as a model-agnostic method, is also compatible with
neural-network architectures other than the MLP backbone used in the main
experiments: Appendix~\ref{appendix:backbone_ablation} pairs naPINN with the
FLS, QRes, PirateNet, and KAN backbones, and shows that the reliability
gate consistently improves over the vanilla baseline across all four
architectures. Training-cost measurements are reported in Appendix~\ref{appendix:training_cost}.

Figure~\ref{fig:visualization} presents qualitative comparisons at two representative time snapshots for all three PDE systems. Standard PINNs exhibit noticeable discrepancies from the ground-truth solutions, particularly under higher outlier ratios, whereas naPINN produces predictions that closely match the true solutions across the spatial domain. The corresponding error maps indicate that naPINN achieves consistently lower reconstruction error compared to the baseline PINN, with reduced sensitivity to localized measurement corruption. This behavior is especially visible in the Burgers' equation: the PINN error is concentrated near regions where the solution changes sharply, while naPINN maintains errors in those regions at a level comparable to smoother parts of the domain. Since the displayed sharp-transition regions are not directly covered by measurement sensors, this improvement suggests that downweighting unreliable noisy data helps the PDE-constrained model learn a better global solution, rather than only cleaning the observed measurements. These visual results qualitatively corroborate the quantitative improvements reported in Table~\ref{table:main_result}.

\subsection{Analysis of noise modeling and gating}

\begin{wrapfigure}[13]{r}{0.48\textwidth}
  \vspace{-2.0\baselineskip}
  \centering
  \includegraphics[width=\linewidth]{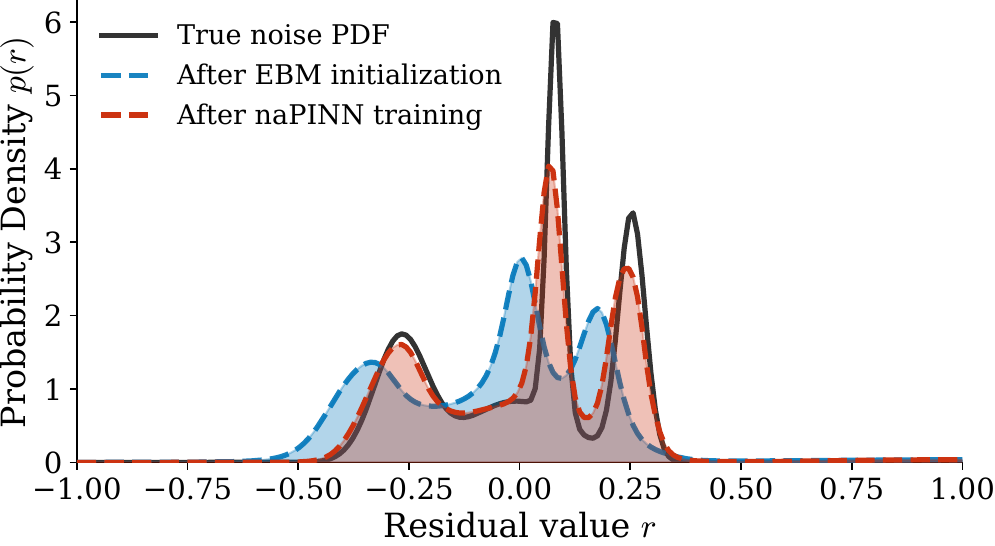}
  \captionsetup{type=figure,skip=0.0pt}
  \caption{Noise density estimated by the EBM before and after joint naPINN training, compared with the injected measurement-noise density.}
  \label{fig:noisecomparison}
\end{wrapfigure}

\paragraph{Noise distribution learning via EBM.} To enable the trainable reliability gate to assess individual measurement points, the EBM estimates the noise distribution from normalized residuals. Through joint optimization of the PINN, EBM, and reliability gate, the components improve one another: the PINN improves the physical prediction, the EBM models the residuals induced by measurement noise, and the gate uses the resulting residual reliability scores to determine whether each data point should influence training.

Figure~\ref{fig:noisecomparison} compares the noise distributions learned by the EBM from residuals at initialization and after the full naPINN training procedure against the ground-truth injected noise density. The results are shown for the Burgers' equation with a 5\% outlier ratio. After joint training, the EBM accurately captures the complex multimodal structure of the four-component Gaussian mixture, whereas it fails to do so at initialization. This behavior indicates that the PINN residuals have become aligned with the intrinsic aleatoric noise profile, while the reliability gate has successfully downweighted gross outliers that are inconsistent with the learned noise distribution.

\begin{wrapfigure}[19]{r}{0.48\textwidth}
  \vspace{-1.0\baselineskip}
  \centering
  {\captionsetup{type=table}
  \caption{Confusion matrix for the trained gate; percentages are over all samples.}
  \label{tab:confusion_matrix}}
  \scriptsize
  \setlength{\tabcolsep}{3pt}
  \renewcommand{\arraystretch}{1.1}
  \begin{tabular}{@{}lcc@{}}
    \toprule
    \textbf{True label} & \textbf{Rejected} & \textbf{Accepted} \\
    \midrule
    \textbf{Outlier} & 2209 (9.8\%$_{\mathrm{TP}}$) & 41 (0.2\%$_{\mathrm{FN}}$) \\
    \textbf{Normal} & 152 (0.7\%$_{\mathrm{FP}}$) & 20098 (89.3\%$_{\mathrm{TN}}$) \\
    \bottomrule
  \end{tabular}
  \par\vspace{1.2\baselineskip}
  \includegraphics[width=\linewidth]{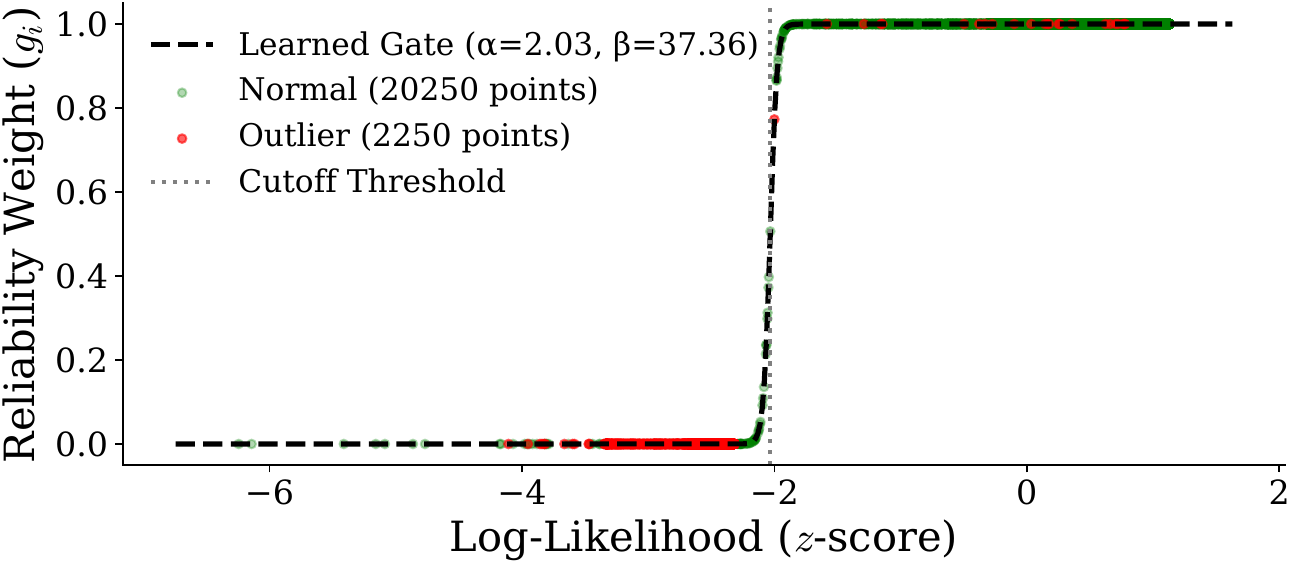}
  \captionsetup{type=figure,skip=1pt}
  \caption{Trained reliability gate for the Allen--Cahn benchmark with a 10\% outlier ratio.}
  \label{fig:sigmoid_analysis}
  \vspace{-0.2\baselineskip}
\end{wrapfigure}

\paragraph{Gating mechanism and classification} To verify that the model explicitly identifies outliers, we visualize the learned gating function (sigmoid) in Figure \ref{fig:sigmoid_analysis} (the 2D Allen-Cahn equation with 10\% of outlier ratio). The plot overlays the probability scores assigned to normal points and outliers. The gating function exhibits a sharp transition, assigning zero weights to most outliers with low residual reliability, while retaining valid data with high residual reliability. Table \ref{tab:confusion_matrix} is a confusion matrix for quantifying the classification performance, which reveals that naPINN achieves near-perfect precision and recall in distinguishing outliers from normal noisy measurements, validating the soft-rejection strategy. Very few normal and outlier points are misclassified, and this is because the interval where outliers are sampled is slightly overlapped with the skewed part of the noise distributions. These results strongly support that naPINN can not only be used as a robust inverse problem solver, but also be applied as an anomaly detection framework for reporting sensor failures or abnormal behavior of system dynamics. Additional analysis of the learned cutoff and steepness parameters is provided in Appendix~\ref{appendix:analysis}.

\section{Conclusion}

We proposed \textbf{naPINN}, a noise-adaptive Physics-Informed Neural Network that robustly recovers physical solutions from corrupted measurements with unknown noise distributions. By estimating residual reliability and using a trainable reliability gate to selectively downweight unreliable data, naPINN mitigates the impact of noise and gross outliers. Experiments on multiple 2D PDE benchmarks demonstrate that naPINN consistently outperforms existing robust PINN variants under severe data corruption, while providing interpretable insight into noise structure and outlier behavior. The main limitation of this study is that the evaluation is based on controlled simulated benchmarks with injected corruptions. While this protocol provides ground-truth fields and known corruption severity for fair stress testing, it does not replace validation on deployment-scale sensor data. Evaluating naPINN on real instrumented systems is an important direction for future work, including settings with spatially correlated sensor failures, temporal drift, and deployment-scale data assimilation.

\clearpage
\bibliographystyle{plainnat}
\bibliography{references}

\medskip


\clearpage
\appendix

\section{Dataset details}
\label{appendix:DatasetDetails}

\subsection{2D Allen-Cahn equation}

We consider the two-dimensional time-dependent Allen--Cahn equation, which typically arises in phase separation and interface dynamics. The formulation of the equation we implemented is as follows:
\begin{equation}
u_t - \varepsilon^2 (\Delta u) + (u^3 - u) = f(x,y,t),
\qquad (x,y) \in \Omega, t \in [t_0, t_1],
\end{equation}
where $\Delta u = u_{xx} + u_{yy}$, $\varepsilon > 0$ controls the interface width, and $f(x,y,t)$ is an external forcing term. We set $\varepsilon = 0.3$ for data generation, but is treated as an unknown parameter during training. It is optimized jointly with the neural network via gradient-based learning, starting from $1.0$, well above the true $\varepsilon=0.3$.

For this equation, we adopt an analytical solution defined as
\begin{equation}
u^* = \sin(\pi x) \sin(\pi y) \cos(\omega t),
\end{equation}
where $\omega$ denotes the temporal frequency. Substituting $u^*$ into the Allen-Cahn operator yields the corresponding forcing term:
\begin{align}
f(x,y,t) &= u^*_t - \varepsilon^2 \Delta u^* + ((u^*)^3-u^*)\\
u^*_t &= -\omega \sin(\pi x) \sin(\pi y) \sin(\omega t), \\
\Delta u^* &= -2 \pi^2 \sin(\pi x) \sin(\pi y) \cos(\omega t).
\end{align}
This construction guarantees that $u^*(x,y,t)$ satisfies the governing PDE exactly.

The spatiotemporal domain is defined as
\begin{equation}
    \Omega = [x_a, x_b] \times [y_a, y_b], \qquad t \in [t_0, t_1],
\end{equation}
where we set $x_a = y_a = t_0 = 0$ and $x_b = y_b = t_1 = 1$ in all experiments.

The initial condition and Dirichlet boundary conditions are imposed using the analytical solution:
\begin{align}
    u(x,y,t_0) &= u^*(x,y,t_0) \\
    u(x,y,t) &= u^*(x,y,t), \qquad (x,y) \in \partial \Omega.
\end{align}

\subsection{2D Burgers' equation}

We consider the two-dimensional incompressible Burgers' equation, a canonical nonlinear convection--diffusion system that serves as a standard benchmark for nonlinear dynamics and turbulence-like behavior. The governing equations are given by
\begin{align}
u_t + u u_x + v u_y &= \nu (u_{xx} + u_{yy}), \\
v_t + u v_x + v v_y &= \nu (v_{xx} + v_{yy}),
\end{align}
where $(u(x,y,t), v(x,y,t))$ denotes the velocity field and $\nu > 0$ is the kinematic viscosity. We fix $\nu=0.01$ during data generation, but treat it as an unknown parameter during training. It is optimized jointly with the neural network via gradient-based learning, starting from $0.0$, the inviscid limit, far from the true $\nu=0.01$.

The spatial domain is defined as
\begin{equation}
\Omega = [x_a, x_b] \times [y_a, y_b],
\end{equation}
with $(x_a, y_a) = (0, 0)$ and $(x_b, y_b) = (4, 4)$.
The temporal interval is $t \in [0, T]$, where a burn-in period of $0.1$ is discarded and the solution is recorded over a duration of $T = 3.0$.

Since no closed-form analytical solution is available, ground-truth data are generated numerically using an explicit finite-difference solver. Spatial derivatives are approximated using second-order central differences, while time integration is performed via an explicit Euler scheme.

The velocity components are initialized using Gaussian random fields with a prescribed power spectrum,
\begin{equation}
    u(x,y,0),\; v(x,y,0) \sim \mathcal{GRF}(\alpha),
\end{equation}
where the spectral decay parameter $\alpha = 5.0$ promotes smooth but spatially diverse initial conditions. This setup produces turbulence-like dynamics often referred to as burgulence.

To ensure numerical stability, the viscous CFL condition is monitored throughout the simulation. After the burn-in phase, solution snapshots are recorded at fixed temporal intervals and used as clean reference data.

\subsection{2D $\lambda-\omega$ reaction--diffusion equation.}

We consider the two-dimensional $\lambda$--$\omega$ reaction--diffusion system, a prototypical nonlinear oscillatory model that exhibits spiral-wave patterns and complex spatiotemporal dynamics. The governing equations are given by
\begin{align}
u_t &= d_u (u_{xx} + u_{yy}) + \lambda(r)\,u - \omega(r)\,v, \\
v_t &= d_v (v_{xx} + v_{yy}) + \omega(r)\,u + \lambda(r)\,v,
\end{align}
where $(u(x,y,t), v(x,y,t))$ denotes the state variables, $d_u, d_v > 0$ are diffusion coefficients, and
\begin{equation}
r^2 = u^2 + v^2.
\end{equation}
The reaction terms are defined as
\begin{equation}
\lambda(r) = 1 - r^2, 
\qquad 
\omega(r) = -\beta r^2,
\end{equation}
with $\beta > 0$ controlling the nonlinear rotation frequency.

The spatial domain is defined as
\begin{equation}
\Omega = [x_a, x_b] \times [y_a, y_b],
\end{equation}
with $(x_a, y_a) = (-10, -10)$ and $(x_b, y_b) = (10, 10)$.
The temporal interval is $t \in [0, T]$ with $T = 10.0$.

For data generation, we fix $d_u = d_v = 1.0$, and $\beta = 1.0$. During training, $\beta$ is treated as an unknown parameter and is optimized jointly with the neural network, starting from $0.0$, far from the true $\beta=1.0$.

Since no closed-form analytical solution is available, ground-truth data are generated numerically using an explicit finite-difference solver. Spatial derivatives are approximated using second-order central differences, while time integration is performed via an explicit Euler scheme.

The system is initialized with a spiral-wave configuration constructed in polar coordinates,
\begin{align}
    u(x,y,0) &= \rho(x,y) \cos(\theta(x,y) - \rho(x,y)), \\
    v(x,y,0) &= \rho(x,y) \sin(\theta(x,y) - \rho(x,y)),
\end{align}
where $\rho(x,y) = \tanh\!\left(\sqrt{x^2 + y^2}\right)$ and $\theta(x,y) = \arg(x + i y)$. This initialization induces rotating spiral patterns that persist throughout the simulation.

Periodic boundary conditions are employed to promote rich spiral dynamics across the domain. Solution snapshots are recorded at fixed temporal intervals and used as clean reference data.

\section{Additional results}
\label{appendix:analysis}

\subsection{Backbone-architecture ablation}
\label{appendix:backbone_ablation}

The proposed reliability-gated framework is conceptually independent of the
specific neural-network family used to parameterize the predictor
$\hat{u}_\theta$. To assess whether this independence holds in practice, we
replace the shared MLP backbone described in
Appendix~\ref{appendix:ImplementationDetails} with four alternative
architectures that have been proposed for physics-informed learning or
general coordinate-based function approximation: FLS, in which the input layer is a learnable sinusoidal feature
mapping designed to enlarge the initial input-gradient
distribution~\cite{Wong2022FLS}; the Quadratic Residual Network (QRes),
which augments each linear layer with a multiplicative residual term
$(W_1 x)\odot(W_2 x)$~\cite{Bu2021QRes}; PirateNet, which combines random
Fourier embeddings with two-step gated residual blocks and a learnable
identity-initialized skip parameter to enable stable deep
training~\cite{Wang2024PirateNet}; and the Kolmogorov--Arnold Network (KAN),
in which each connection is parameterized by a learnable univariate
B-spline activation~\cite{Liu2024KAN}. The implementations follow the
authors' official reference code line-by-line; full specifications are
deferred to Appendix~\ref{appendix:backbone_implementations}.

For each backbone we run two settings while keeping every other component
of the training pipeline fixed: a \emph{Vanilla} run using the standard
PDE-residual and MSE data loss in Eq.~\eqref{eq:standard_pinn}, and a
\emph{naPINN} run using the reliability-gated objective in
Eq.~\eqref{eq:joint_pinn_gate} with the EBM residual estimator and the same
staged warm-up schedule used in the main experiments. All experiments use a
fixed $10\%$ outlier ratio and otherwise follow the data-generation protocol
described in Section~\ref{sec:experiments}; results are averaged over $N$
independent trials with different random seeds. We do not match the parameter
counts across backbones because the purpose of this ablation is to verify
that the gains from reliability-gated training transfer across architectural
families, rather than to compare the raw expressiveness of the backbones; the
parameter counts of each model are listed in
Appendix~\ref{appendix:backbone_implementations} for transparency.

\begin{sc}
\begin{table}[t]
\centering
\small
\caption{Backbone-architecture ablation under 10\% outlier corruption.
We pair each backbone with both the vanilla PINN training objective
(PDE residual + MSE data loss) and the proposed naPINN reliability-gated
training, keeping all other hyperparameters fixed within each backbone.
Results are averaged over $N$ independent trials with different random seeds.
Lower rMAE and rMSE indicate better reconstruction.
The MLP rows correspond to the configuration used in the main results
(Table~\ref{table:main_result}).
For each benchmark, the lowest error across all backbone--method pairs is
\textbf{bold} and the second lowest is \underline{underlined}.}
\label{tab:backbone_comparison}
\setlength{\tabcolsep}{4pt}
\begin{tabular}{llcccccc}
\toprule
& & \multicolumn{2}{c}{Allen--Cahn} & \multicolumn{2}{c}{Burgers} & \multicolumn{2}{c}{$\lambda$--$\omega$ RD} \\
\cmidrule(lr){3-4}\cmidrule(lr){5-6}\cmidrule(lr){7-8}
Backbone & Method & rMAE & rMSE & rMAE & rMSE & rMAE & rMSE \\
\midrule

\multirow{2}{*}{MLP}
 & Vanilla & 0.584 & 0.488 & 0.387 & 0.377 & 0.273 & 0.292 \\
 & naPINN  & 0.110 & 0.108 & \textbf{0.074} & 0.092 & 0.074 & 0.092 \\
\cmidrule(lr){1-8}

\multirow{2}{*}{KAN~\cite{Liu2024KAN}}
 & Vanilla & 0.564 & 0.473 & 0.397 & 0.393 & 0.269 & 0.293 \\
 & naPINN  & 0.110 & 0.105 & 0.085 & \underline{0.091} & \underline{0.055} & 0.069 \\
\cmidrule(lr){1-8}

\multirow{2}{*}{FLS~\cite{Wong2022FLS}}
 & Vanilla & 0.617 & 0.513 & 0.417 & 0.403 & 0.259 & 0.285 \\
 & naPINN  & \textbf{0.080} & \textbf{0.073} & \underline{0.080} & \textbf{0.088} & \underline{0.055} & \underline{0.068} \\
\cmidrule(lr){1-8}

\multirow{2}{*}{QRes~\cite{Bu2021QRes}}
 & Vanilla & 0.593 & 0.497 & 0.394 & 0.388 & 0.240 & 0.269 \\
 & naPINN  & \underline{0.103} & \underline{0.097} & 0.085 & 0.092 & 0.057 & 0.069 \\
\cmidrule(lr){1-8}

\multirow{2}{*}{PirateNet~\cite{Wang2024PirateNet}}
 & Vanilla & 0.615 & 0.522 & 0.400 & 0.396 & 0.253 & 0.278 \\
 & naPINN  & 0.113 & 0.108 & 0.079 & \textbf{0.088} & \textbf{0.046} & \textbf{0.056} \\

\bottomrule
\end{tabular}
\end{table}
\end{sc}

Table~\ref{tab:backbone_comparison} reports rMAE and rMSE on the three PDE
benchmarks. Across every backbone, naPINN reduces both error metrics
relative to its vanilla counterpart, mirroring the trend observed for the MLP
backbone in the main results table. 
The relative gain from reliability gating is comparable in magnitude across
the four alternative backbones, indicating that the benefit of suppressing
unreliable measurements is largely orthogonal to the choice of network
architecture rather than tied to a specific function-class bias.
These results support the use of naPINN as a modular reliability-gated
training procedure that can be paired with the practitioner's preferred PINN
architecture.

\subsection{Additional noise distributions}
\label{appendix:additional_noise_distributions}

The main experiments use a multimodal Gaussian-mixture noise distribution with gross outliers.
To further examine whether the proposed reliability-gated framework depends on this particular corruption model, we additionally evaluate Laplace and Gaussian noise settings on the 2D $\lambda$--$\omega$ reaction--diffusion benchmark with a fixed outlier ratio of 15\%.
These cases are useful stress tests because the standard MSE loss corresponds to a Gaussian likelihood, while the $\ell_1$ loss used by LAD-PINN corresponds to a Laplace likelihood.
If matching the nominal likelihood family were sufficient, PINN under Gaussian noise and LAD-PINN under Laplace noise would be expected to perform well.

\begin{sc}
\begin{table}[h]
\centering
\small
\caption{Additional results under Laplace and Gaussian measurement noise on the 2D $\lambda$--$\omega$ reaction--diffusion benchmark with 15\% outliers. Values report rMSE; lower is better.}
\label{tab:additional_noise_distributions}
\setlength{\tabcolsep}{8pt}
\begin{tabular}{lcc}
\toprule
Method & Laplace & Gaussian \\
\midrule
PINN & 0.323 & 0.326 \\
LAD-PINN & 0.085 & 0.109 \\
OrPINN ($q=2.9$) & 0.081 & 0.107 \\
naPINN & 0.085 & 0.126 \\
naPINN + L1 & \textbf{0.059} & \textbf{0.064} \\
naPINN + $q$-Gaussian ($q=2.9$) & 0.071 & 0.067 \\
\bottomrule
\end{tabular}
\end{table}
\end{sc}

Table~\ref{tab:additional_noise_distributions} shows that likelihood-matched fixed losses alone do not fully address corrupted measurement learning.
Although LAD-PINN and OrPINN substantially improve over the standard PINN, their errors remain higher than those obtained by combining robust losses with naPINN's reliability gate.
The naPINN variant using MSE achieves performance comparable to the robust-loss baselines, indicating that adaptive reliability gating already removes much of the harmful influence of unreliable measurement data.
When the data are sufficiently noisy or heavy-tailed, the framework can also be combined with a robust data loss: naPINN + L1 gives the best results in both additional settings, and naPINN + $q$-Gaussian also improves over the corresponding robust baseline.
These results support using naPINN as a flexible reliability-gated framework whose residual estimator and data loss can be chosen jointly based on the expected data characteristics.

\subsection{Gate parameter optimization}

\begin{figure}[h]
  \centering
  \includegraphics[width=\linewidth]{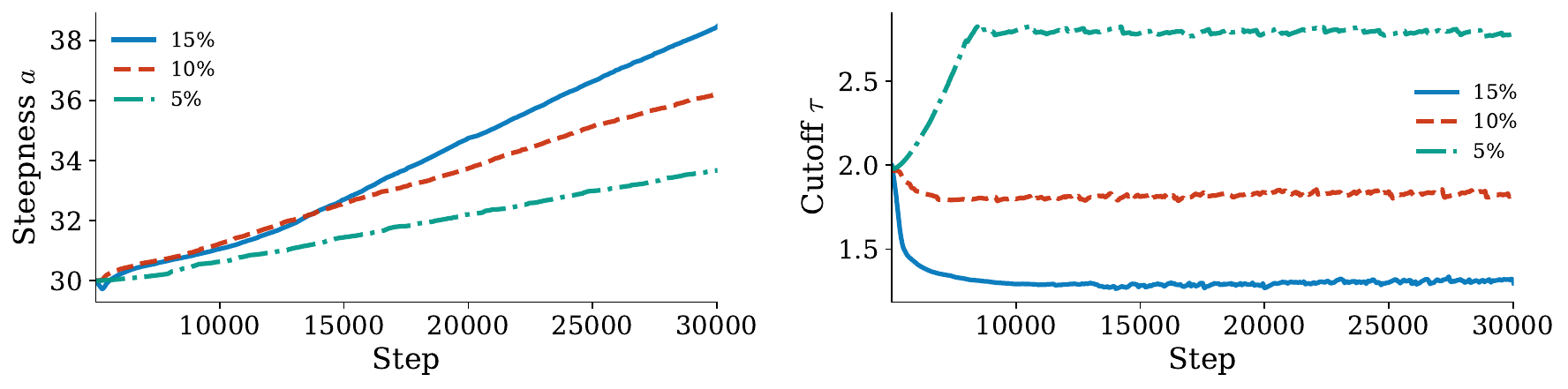}
  \caption{Evolution of the reliability gate parameters during training: steepness parameter $a$ (left) and cutoff parameter $\tau$ (right).}
  \label{fig:gate_params}
\end{figure}

Figure~\ref{fig:gate_params} illustrates the evolution of the reliability gate parameters--the cutoff parameter $\tau$ and the steepness parameter $a$--during training for different outlier ratios in the $\lambda$--$\omega$ reaction--diffusion experiment. As the outlier ratio increases, the learned cutoff $\tau$ shifts toward lower score values, indicating that the gate becomes more conservative by rejecting (filtering out) residuals that need not be extremely anomalous to be treated as unreliable. This behavior demonstrates that the reliability gate adaptively learns which measurement data points should be regarded as outliers and selectively filters them to improve overall model performance. Also, in all experiments, the cutoff parameter $\tau$ exhibits a quick adjustment during the early steps of training, and then remains constant throughout the rest of the steps. This stability implies that the reliability gate effectively calibrates the decision boundary between reliable data points and outliers early in training.

The steepness parameter $a$ consistently increases over the course of training. This trend suggests that the gate progressively enforces a sharper separation between normal data points and outliers, largely independent of the exact score values. Such a sharp transition implies that, once identified as reliable, normal data points are weighted nearly uniformly during training, which is desirable for stable optimization and robust learning.

\subsection{PDE parameter reconstruction}
\label{appendix:parameter_reconstruction}

Beyond reconstructing the solution field, our inverse-problem setup also identifies the unknown PDE coefficient via gradient-based joint optimization with the network. While the field-level metrics (rMAE, rMSE) reported in Table~\ref{table:main_result} are our primary measure of reconstruction quality, the recovered coefficient provides a complementary view of physical fidelity: a method that fits noisy measurements at the expense of a biased coefficient would reproduce the observed field but misidentify the underlying dynamics.

We examine parameter recovery on the 2D Allen--Cahn benchmark, where the interface-width parameter $\varepsilon$ is treated as unknown during training, initialized at $\varepsilon_{0}=1.0$, and optimized jointly with the network toward the ground-truth value $\varepsilon^{\star}=0.3$ used for data generation (Appendix~\ref{appendix:DatasetDetails}). Table~\ref{tab:parameter_reconstruction} reports the final estimate $\hat{\varepsilon}$ and the absolute parameter error $|\hat{\varepsilon}-\varepsilon^{\star}|$ for the standard PINN and naPINN at the three outlier ratios used in the main experiments.

\begin{sc}
\begin{table}[h]
\centering
\small
\caption{PDE parameter reconstruction on the 2D Allen--Cahn benchmark. The interface-width parameter $\varepsilon$ is treated as unknown and optimized jointly with the network from initialization $\varepsilon_{0}=1.0$ toward the ground-truth value $\varepsilon^{\star}=0.3$. We report the final estimate $\hat{\varepsilon}$ together with the absolute parameter error $|\hat{\varepsilon}-\varepsilon^{\star}|$ at three outlier ratios; smaller absolute error indicates more accurate parameter recovery. The lowest absolute error in each column is in \textbf{bold}.}
\label{tab:parameter_reconstruction}
\setlength{\tabcolsep}{5pt}
\begin{tabular}{lcccccc}
\toprule
& \multicolumn{2}{c}{5\% outliers} & \multicolumn{2}{c}{10\% outliers} & \multicolumn{2}{c}{15\% outliers} \\
\cmidrule(lr){2-3}\cmidrule(lr){4-5}\cmidrule(lr){6-7}
Method & $\hat{\varepsilon}$ & $|\hat{\varepsilon}-\varepsilon^{\star}|$ & $\hat{\varepsilon}$ & $|\hat{\varepsilon}-\varepsilon^{\star}|$ & $\hat{\varepsilon}$ & $|\hat{\varepsilon}-\varepsilon^{\star}|$ \\
\midrule
Vanilla PINN & 0.2914 & 0.0087 & 0.2892 & 0.0108 & 0.2903 & 0.0097 \\
naPINN       & \textbf{0.2980} & \textbf{0.0021} & \textbf{0.2981} & \textbf{0.0019} & \textbf{0.2959} & \textbf{0.0041} \\
\bottomrule
\end{tabular}%
\end{table}
\end{sc}

Both methods recover $\varepsilon$ in the correct neighborhood of the true value, indicating that joint estimation of the network and the PDE coefficient remains well-posed even under corrupted measurements. However, naPINN produces estimates substantially closer to $\varepsilon^{\star}=0.3$ than the vanilla baseline across all three outlier ratios, reducing the absolute parameter error by roughly $2$--$6\times$. The vanilla PINN systematically underestimates $\varepsilon$ by about $0.01$ at every outlier level, indicating that gross outliers introduce a persistent bias when they enter the squared data loss and propagate through the gradient flowing into the PDE coefficient. naPINN keeps this bias well below $0.005$ in all cases. This trend is consistent with the field-level results in Table~\ref{table:main_result}: by downweighting unreliable measurements, the reliability gate not only improves solution-field reconstruction but also prevents corrupted observations from biasing the gradient signal driving the unknown PDE coefficient, yielding a more faithful identification of the underlying physics.

\subsection{Training cost}
\label{appendix:training_cost}

Table~\ref{tab:training_cost} compares the training cost of a standard PINN and naPINN on the three benchmarks with 15\% outliers.
These naPINN runs use the EBM estimator, which is the most computationally expensive estimator component considered in this work.
Therefore, the overhead reported here is a conservative estimate for the modular framework; KDE or GMM estimator variants are expected to incur smaller estimator cost.
All timings are measured in seconds per training epoch on a single NVIDIA RTX A6000 GPU.

\begin{sc}
\begin{table}[t]
\centering
\small
\caption{Training cost comparison between PINN and naPINN under 15\% outlier corruption. Values are measured in seconds per training epoch on a single NVIDIA RTX A6000 GPU. The naPINN results use the EBM estimator. The parenthesized values report the relative increase over PINN.}
\label{tab:training_cost}
\setlength{\tabcolsep}{8pt}
\begin{tabular}{lcc}
\toprule
Benchmark & PINN (s/epoch) & naPINN (EBM; s/epoch) \\
\midrule
Allen--Cahn & 0.0356 & 0.0397 (+10.3\%) \\
Burgers & 0.0423 & 0.0478 (+11.5\%) \\
$\lambda$--$\omega$ RD & 0.0458 & 0.0472 (+3.0\%) \\
\bottomrule
\end{tabular}%
\end{table}
\end{sc}

The additional cost of naPINN is modest despite the extra estimator and reliability-gate updates.
Across the three benchmarks, the relative overhead ranges from 3.0\% to 11.5\% when using the EBM estimator.
This suggests that the reliability-gated training procedure improves robustness without substantially increasing the computational budget.

\subsection{Supplementary material}
Additional qualitative results, including videos of the predicted solution evolution, are provided in the supplementary material with code.

\section{Ablation study}
\label{appendix:ablation}

\subsection{Estimator module comparison}
\label{appendix:estimator_comparison}

\begin{sc}
\begin{table*}
\centering
\small
\caption{Prediction performance of naPINN with different residual-based noise distribution estimators. Lower rMSE and rMAE indicate better reconstruction. The results show that KDE, GMM, and EBM yield broadly comparable prediction accuracy, supporting the modularity of the residual reliability estimator.}
\label{tab:estimator_comparison}
\setlength{\tabcolsep}{4pt}
\begin{tabular}{llcccccc}
\toprule
 & & \multicolumn{2}{c}{5\% outliers} & \multicolumn{2}{c}{10\% outliers} & \multicolumn{2}{c}{15\% outliers} \\
\cmidrule(lr){3-4}\cmidrule(lr){5-6}\cmidrule(lr){7-8}
Benchmark & Estimator & rMSE & rMAE & rMSE & rMAE & rMSE & rMAE \\
\midrule
\multirow{3}{*}{Allen--Cahn}
 & EBM & 0.101 & 0.104 & 0.108 & 0.110 & 0.127 & 0.134 \\
 & GMM & 0.110 & 0.110 & 0.082 & 0.081 & 0.113 & 0.117 \\
 & KDE & 0.121 & 0.132 & 0.110 & 0.117 & 0.091 & 0.090 \\
\midrule
\multirow{3}{*}{Burgers}
 & EBM & 0.089 & 0.075 & 0.092 & 0.074 & 0.091 & 0.072 \\
 & GMM & 0.101 & 0.083 & 0.103 & 0.085 & 0.106 & 0.093 \\
 & KDE & 0.096 & 0.086 & 0.109 & 0.104 & 0.109 & 0.101 \\
\midrule
\multirow{3}{*}{$\lambda$--$\omega$ RD}
 & EBM & 0.092 & 0.073 & 0.092 & 0.074 & 0.095 & 0.076 \\
 & GMM & 0.092 & 0.078 & 0.096 & 0.081 & 0.096 & 0.082 \\
 & KDE & 0.100 & 0.083 & 0.105 & 0.090 & 0.100 & 0.083 \\
\bottomrule
\end{tabular}%
\end{table*}
\end{sc}

\begin{figure}
  \centering
    \includegraphics[width=0.6\linewidth]{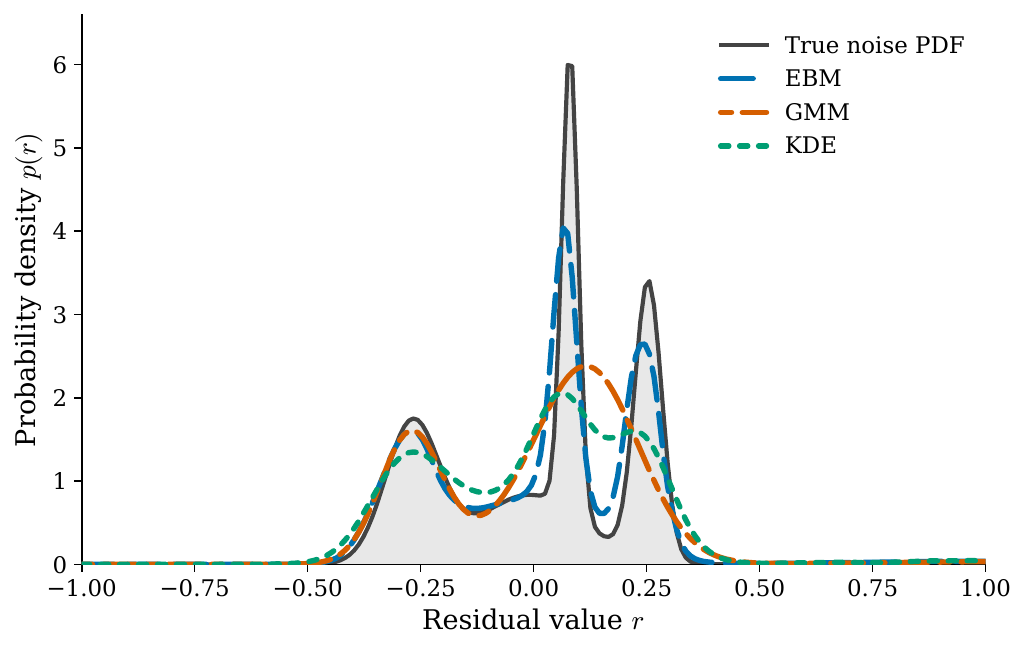}%
    \caption{Comparison of the noise distributions estimated by EBM, KDE, and GMM from normalized residuals. The estimator implementations and hyperparameters are described in Appendix~\ref{appendix:ImplementationDetails}.}
    \label{fig:estimator_noise_distribution_comparison}
\end{figure}

naPINN is formulated around a residual-based noise distribution estimator rather than a particular density model.
To evaluate this modularity, we replace the EBM estimator used in the main experiments with KDE and GMM estimators while keeping the same reliability-gated training pipeline.
Table~\ref{tab:estimator_comparison} reports prediction performance across all three PDE benchmarks and outlier ratios, and Figure~\ref{fig:estimator_noise_distribution_comparison} compares the corresponding noise-distribution estimates.
All three estimators use the same normalized-residual interface and trainable reliability gate; their implementation details are reported separately in Appendix~\ref{appendix:ImplementationDetails}.

The estimator choice does not fundamentally change the downstream reconstruction behavior.
KDE and GMM sometimes match or exceed the EBM in prediction error, most notably on Allen--Cahn, while EBM is consistently strongest on the Burgers and $\lambda$--$\omega$ RD benchmarks.
These module-swap results indicate that naPINN is not restricted to a particular density estimator; instead, it provides a flexible and robust reliability-gated framework in which different residual-based estimators can be plugged into the same inverse-PINN training pipeline.
The noise-distribution estimates provide a complementary view.
EBM most closely approximates the injected noise distribution, especially when the residual landscape is multimodal or otherwise structurally complex.
KDE also reflects the broad shape of the noise landscape and captures more of its nonparametric structure than the GMM in this setting, although it is less accurate than the EBM.
When the underlying noise distribution is simple, KDE and GMM can provide sufficiently similar approximations; however, in practical inverse problems the noise distribution is typically unknown before training.
We therefore use EBM as the default component because it offers a flexible trainable density model and gives the most faithful estimate of the noise structure in this comparison, while KDE and GMM remain viable plug-in alternatives when their assumptions or computational profile are preferred.

\subsection{Staged training}
\label{appendix:staged_training}

\begin{sc}
\begin{table}
\centering
\small
\caption{Ablation on staged training for naPINN under 15\% outlier corruption. Results are averaged over 5 independent trials with different random seeds, with standard deviations shown in parentheses.}
\label{tab:ablation_staged_training}
\setlength{\tabcolsep}{4pt}
\begin{tabular}{llccc}
\toprule
 & & \multicolumn{1}{c}{Allen--Cahn} & \multicolumn{1}{c}{Burgers} & \multicolumn{1}{c}{$\lambda$--$\omega$ RD} \\
Method & Metric & 15\% & 15\% & 15\% \\
\midrule

\multirow{2}{*}{\textbf{naPINN (w/ staged training)}}
 & rMAE & \pmstd{0.134}{0.060} & \pmstd{0.072}{0.014} & \pmstd{0.076}{0.006} \\
 & rMSE & \pmstd{0.127}{0.049} & \pmstd{0.091}{0.006} & \pmstd{0.095}{0.006} \\
\cmidrule(lr){2-5}

\multirow{2}{*}{\textbf{naPINN (w/o staged training)}}
 & rMAE & \pmstd{0.638}{0.304} & \pmstd{0.074}{0.005} & \pmstd{0.075}{0.002} \\
 & rMSE & \pmstd{0.532}{0.242} & \pmstd{0.098}{0.010} & \pmstd{0.098}{0.005} \\
\bottomrule
\end{tabular}%
\end{table}
\end{sc}

We conduct an ablation study to examine the role of staged training in naPINN.
Specifically, we compare the full naPINN framework with staged training against a variant trained without warm-up or explicit stage separation, where the PINN, energy-based model (EBM), and reliability gate are optimized jointly from initialization.

Experiments are performed on three PDE benchmarks (Allen--Cahn 2D, Burgers 2D, and $\lambda$--$\omega$ reaction--diffusion) under a fixed outlier ratio of 15\%.
For each benchmark, relative mean absolute error (rMAE) and relative root mean squared error (rMSE) are reported, averaged over five independent runs with different random seeds.
The quantitative results are summarized in Table~\ref{tab:ablation_staged_training}.

Overall, naPINN remains trainable and achieves competitive performance even without staged training.
However, the impact of removing staged training varies across benchmarks.
For the Allen--Cahn equation, training without warm-up leads to a significant degradation in performance, indicating that premature coupling between the PINN residuals and an insufficiently trained EBM can hinder stable optimization.
In contrast, for the Burgers and $\lambda$--$\omega$ reaction--diffusion equations, the performance gap between staged and non-staged training is relatively small, with non-staged training yielding slightly worse but comparable results.

These observations suggest that while staged training is not strictly necessary for convergence in all cases, it plays a crucial role in stabilizing optimization for more challenging or sensitive PDE systems.
In particular, staged training allows the PINN to first establish a reasonable solution estimate before the EBM and reliability gate actively influence the learning dynamics, thereby preventing early mischaracterization of residuals as noise or outliers.

\subsection{Robustness to rejection cost}
\label{appendix:rejection_cost}

\begin{figure}
  \centering
    \includegraphics[width=1\textwidth]{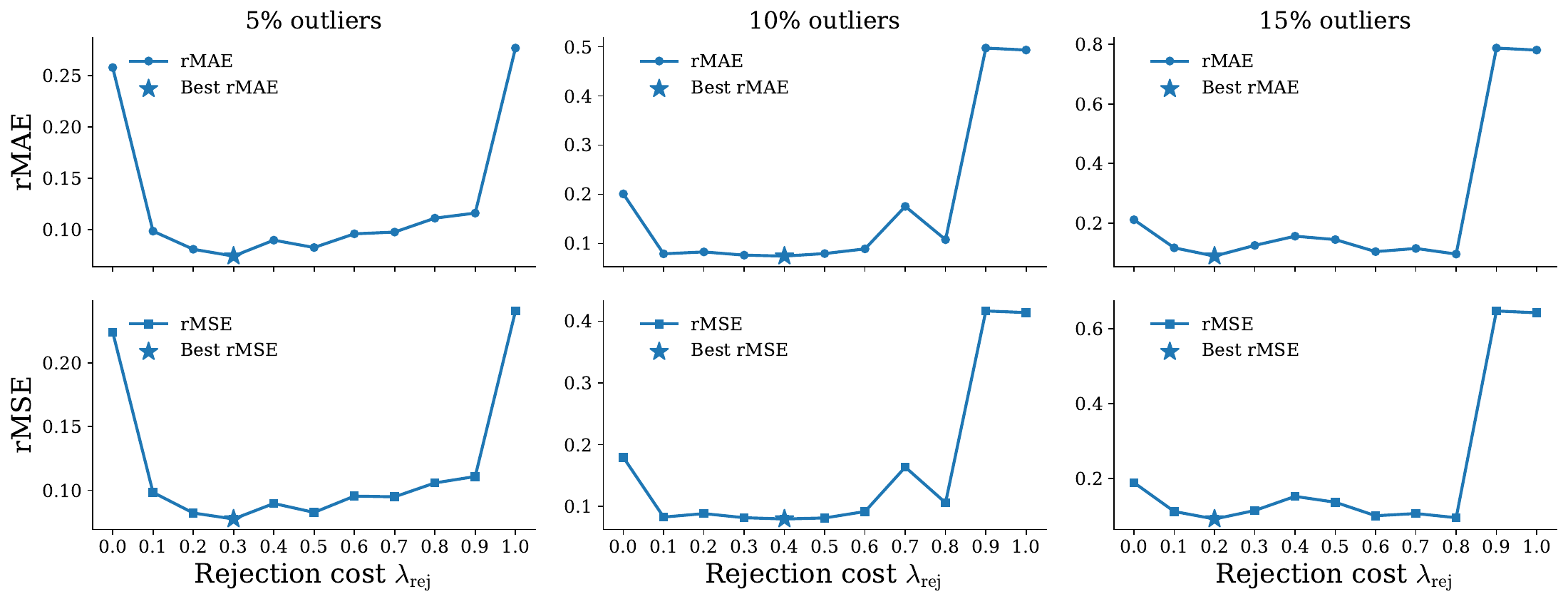}
    \caption{Sensitivity to the rejection-cost coefficient on Allen--Cahn under 5\%, 10\%, and 15\% outlier ratios from left to right. We sweep $\lambda_{\mathrm{rej}}$ and report rMAE and rMSE of naPINN. Across all three corruption levels, performance remains stable for $\lambda_{\mathrm{rej}}\in[0.1,0.8]$. Removing the rejection cost ($\lambda_{\mathrm{rej}}=0$) increases the error, while overly large coefficients ($0.9$ or $1.0$) degrade performance.}
    \label{fig:rej_cost_sensitivity}
\end{figure}

We further investigate the sensitivity of naPINN to the rejection cost $\lambda_{\mathrm{rej}}$, which regularizes the trainable reliability gate and prevents the trivial solution where all measurement points are rejected to artificially reduce the training loss.
We train naPINN on the Allen--Cahn equation with outlier ratios of 5\%, 10\%, and 15\%, sweeping $\lambda_{\mathrm{rej}} \in \{0, 0.1, 0.2, \ldots, 1.0\}$ while keeping all other settings unchanged.

Figure~\ref{fig:rej_cost_sensitivity} summarizes the results.
Importantly, selecting an appropriate scale for $\lambda_{\mathrm{rej}}$ is not difficult in practice, as the typical magnitude of the per-point data loss is known.
Since the rejection cost directly competes with the data loss, it suffices to set $\lambda_{\mathrm{rej}}$ smaller than the characteristic data loss magnitude to effectively regularize the gate and prevent degenerate solutions.
Within this regime, reducing $\lambda_{\mathrm{rej}}$ does not harm training stability or reconstruction accuracy.

Consistent with this interpretation, naPINN exhibits stable performance across a broad range of rejection-cost values.
Across all three outlier ratios, the error curves remain nearly flat for $\lambda_{\mathrm{rej}}\in[0.1,0.8]$, indicating that the method is not sensitive to fine-grained tuning of this coefficient.
The $\lambda_{\mathrm{rej}}=0$ case, which removes the rejection-cost term, increases the reconstruction error because the gate can suppress too many measurements without penalty.
This degradation confirms that the rejection cost is necessary for avoiding degenerate reliability gating.

In contrast, excessively large rejection costs ($\lambda_{\mathrm{rej}}=0.9$ or $1.0$) substantially degrade performance.
In this regime, rejecting measurement points becomes overly penalized, effectively disabling the selective learning mechanism and forcing the model to fit corrupted observations.
These results confirm that naPINN is robust to the choice of $\lambda_{\mathrm{rej}}$ over a wide intermediate range, while also showing the expected failure modes when the rejection cost is absent or too large.

\subsection{Running standard deviation normalization}
\label{appendix:running_std_ablation}

The default naPINN implementation normalizes residuals using a running standard deviation updated by EMA.
This avoids recomputing the normalization scale solely from the current minibatch, whose empirical standard deviation may be distorted by gross outliers.
Table~\ref{tab:running_std_ablation} compares this default against a per-batch variant that uses the standard deviation of each minibatch without EMA smoothing.

\begin{sc}
\begin{table}[t]
\centering
\small
\caption{Ablation of residual normalization for naPINN. The EMA variant uses a running standard deviation, while the per-batch variant recomputes the scale from each minibatch without EMA smoothing. Values are rMSE; lower is better.}
\label{tab:running_std_ablation}
\setlength{\tabcolsep}{5pt}
\begin{tabular}{llccc}
\toprule
Benchmark & Normalization & 5\% & 10\% & 15\% \\
\midrule
\multirow{2}{*}{Allen--Cahn}
 & EMA & 0.101 & 0.108 & 0.127 \\
 & Per-batch & 0.182 & 0.141 & 0.200 \\
\midrule
\multirow{2}{*}{Burgers}
 & EMA & 0.089 & 0.092 & 0.091 \\
 & Per-batch & 0.086 & 0.101 & 0.111 \\
\midrule
\multirow{2}{*}{$\lambda$--$\omega$ RD}
 & EMA & 0.092 & 0.092 & 0.095 \\
 & Per-batch & 0.105 & 0.096 & 0.098 \\
\bottomrule
\end{tabular}%

\end{table}
\end{sc}

EMA normalization improves stability in most settings and prevents severe failures under difficult corruption regimes.
The effect is clearest for Allen--Cahn across all outlier ratios where per-batch normalization becomes unstable.

\section{Implementation details}
\label{appendix:ImplementationDetails}

\subsection{Model implementation}
All methods used in this paper share the same MLP backbone that maps spatiotemporal coordinates to the target physical state:
\begin{equation}
    \mathcal{N}_\theta: \mathbb{R}^{d_\mathrm{in}} \rightarrow \mathbb{R}^{d_\mathrm{out}},
\end{equation}
where the input dimension $d_\mathrm{in}$ corresponds to the spatiotemporal coordinates $(x,y,t)$, and the output dimension $d_\mathrm{out}$ corresponds to the target physical variables (e.g., $u$, $(u, v)$).

The MLP consists of $H=5$ hidden layers, hidden width $W=80$ for all layers, hyperbolic tangent ($\tanh$) activation functions, and a final linear output layer. All linear layers are initialized using Xavier uniform initialization, with biases initialized to zero. This architecture is fixed across all experiments to ensure a fair comparison between baselines and naPINN.

\subsection{Residual-based noise estimator modules}
All estimator variants in the ablation study use the same residual interface.
For a measurement residual $r_i=y_i-\hat{u}_\theta(x_i,t_i)$, we first divide by the running standard deviation described in Sec.~\ref{subsec:stage1}.
For multi-output systems such as Burgers and $\lambda$--$\omega$ RD, residual components are flattened and scored as scalar residual samples.
The estimator then returns a scalar log-density or equivalent reliability score for each normalized residual, and the same trainable reliability gate and rejection-cost term are used across all estimator variants.
Thus, the KDE, GMM, and EBM ablations differ only in the residual-density estimator, while the PINN architecture, optimizer, training schedule, data batches, residual normalization, and gate objective are kept fixed.

\paragraph{Energy-based estimator.}
The EBM used in the main experiments is implemented as a lightweight MLP that maps a normalized residual value $r$ to an unnormalized log-density $\log q_\phi (r)$:
\begin{equation}
    \mathcal{N}_\phi: \mathbb{R} \rightarrow \mathbb{R},
\end{equation}
The MLP has three hidden layers of width $32$ with hyperbolic tangent ($\tanh$) activations, followed by a scalar linear output.
The EBM is optimized with Adam using learning rate $10^{-3}$.
Its normalization constant is approximated by trapezoidal integration over a fixed one-dimensional grid on $[-10,10]$; the grid uses $1024$ points for all three benchmarks. The estimator is initialized for $5{,}000$ iterations after PINN warm-up and then updated during joint training on the current normalized residuals.

\paragraph{Mixture and kernel estimators.}
The GMM and KDE variants replace the EBM with alternative density estimators on the same normalized residual samples.
The GMM variant uses a trainable diagonal-covariance Gaussian mixture, implemented with mixture logits, component means, and component log standard deviations as learnable parameters.
Unless otherwise stated, it uses three mixture components, initializes the component means near zero, and minimizes the residual negative log likelihood with Adam using learning rate $10^{-2}$.
The KDE variant is nonparametric: it maintains a buffer of up to $2048$ recent normalized residual samples and computes Gaussian-kernel log densities against this buffer.
Its bandwidth is chosen by Silverman's rule of thumb,
\begin{equation}
h = 0.9 \min\!\left(s,\frac{\mathrm{IQR}}{1.34}\right)n^{-1/5},
\end{equation}
with a small lower clamp for numerical stability.
Because KDE has no learnable parameters, its training step refreshes the residual buffer and bandwidth rather than applying gradient updates.
For the GMM variant, the gate uses the current trainable mixture log density.
For the KDE variant, it uses the current kernel log density.
When these quantities are discussed as reliability scores, the sign is chosen so that larger scores indicate lower residual reliability, matching the convention in the Method section.
During the module-swap ablation, these estimators are updated or refreshed on residual samples from the current PINN and then passed through the same gate used by the EBM variant.
This keeps the ablation focused on the residual-density estimator rather than on changes to the downstream weighting mechanism.

\paragraph{Trainable reliability gate.}
The gate parameters $(a, \tau)$ in Eq.~\eqref{eq:gate} are constrained to be
positive via softplus reparameterization; the pre-softplus values are the
actual learnable variables, with the cutoff initialized so that $\tau$
starts roughly two standard deviations below the mean score to prevent
early collapse to mass rejection. For density estimators whose native output
is a log-density $\log p_\phi$ rather than the score $s_\phi$ used in the
main text, our implementation evaluates the equivalent form
\begin{equation}
g_i = \sigma\!\big(a\,(\tilde z_i + \tau)\big), \qquad
\tilde z_i := \frac{\log p_\phi(\tilde r_i) - \mu_{\log p}}{\sigma_{\log p}},
\label{eq:gate_impl}
\end{equation}
where $\mu_{\log p}, \sigma_{\log p}$ are the minibatch mean and standard
deviation of $\log p_\phi$. Eq.~\eqref{eq:gate_impl} is identical to
Eq.~\eqref{eq:gate} under $s_\phi = -\log p_\phi$, but avoids an explicit
sign flip in the estimator interface. The rejection regularizer in
Eq.~\eqref{eq:rej_cost} is averaged over the minibatch, with
$\lambda_{\mathrm{rej}}=1.0$ for Allen--Cahn and $\lambda_{\mathrm{rej}}=0.5$
for the Burgers and $\lambda$--$\omega$ RD benchmarks in the reported runs.
The estimator module is selected via \texttt{density\_estimator} with
options \texttt{ebm}, \texttt{gmm}, and \texttt{kde}; residual scaling uses
either the default EMA running standard deviation or a per-batch standard
deviation for the normalization ablation in
Appendix~\ref{appendix:running_std_ablation}.

\subsection{Training hyperparameters}
\label{appendix:training_hp}
All methods use the Adam optimizer with learning rate $10^{-3}$ for the PINN
backbone and unknown PDE parameters; no learning-rate scheduler is applied.
One training step uses a minibatch of $N_c$ collocation points and $N_d$
measurement points sampled with replacement from $\mathcal{D}_c, \mathcal{D}_d$.
The total budget is $K_{\text{tot}}=30{,}000$ steps for all methods.
For naPINN, this consists of $K_w = 5{,}000$ warm-up steps followed by
$25{,}000$ joint-training steps; an additional $K_{\text{est}}=5{,}000$
estimator-initialization steps update only $\phi$ and are not counted toward
$K_{\text{tot}}$. The EMA decay constant in Eq.~\eqref{eq:ema_std} is
$\rho = 0.99$, with the empirical standard deviation clipped from below at
$10^{-6}$ for numerical stability.
The gate parameters $(a, \tau)$ share the PINN learning rate.
The B-PINN baseline draws $K_{\text{MC}}=1$ weight sample per training step
via the reparameterization trick, and uses the posterior mean (no MC averaging)
at evaluation time.

\subsection{Alternative backbones}
\label{appendix:backbone_implementations}

For the backbone ablation in
Appendix~\ref{appendix:backbone_ablation}, we replace the shared MLP backbone
of Appendix~\ref{appendix:ImplementationDetails} with four alternative
architectures. To minimize implementation drift, every backbone is reproduced
from the authors' official reference code, and we follow the layer equations
as written in those references rather than introducing custom variants.
Each backbone consumes the same spatiotemporal input
$(x,y,t)\in\mathbb{R}^{d_\mathrm{in}}$ and produces the same target physical
state in $\mathbb{R}^{d_\mathrm{out}}$ as the shared MLP, and uses a final
linear output head. Unless otherwise stated, each backbone uses $H=5$ layers
(or residual blocks) and width $W=80$, matching the MLP. We do not match the
total parameter counts across backbones; the parameter count of each backbone
at this configuration is summarized in Table~\ref{tab:backbone_params} for
transparency.

\paragraph{FLS (sinusoidal feature mapping).}
FLS~\cite{Wong2022FLS} replaces the input layer of the MLP with a
learnable sinusoidal feature mapping designed to enlarge the initial input
gradient distribution. Let $W_1\in\mathbb{R}^{n\times d_\mathrm{in}}$ and
$b_1\in\mathbb{R}^{n}$. The first layer computes
\begin{equation}
    \gamma(x) \;=\; \sin\!\big(2\pi (W_1 x + b_1)\big),
\end{equation}
where the entries of $W_1$ are initialized i.i.d.\ as
$\mathcal{N}(0,\sigma_F^{2})$ and $b_1$ is initialized to zero. Both $W_1$
and $b_1$ are trainable. The $n$-dimensional features $\gamma(x)$ are then
passed through $H-1$ standard $\tanh$ hidden layers of width $W$ and a final
linear head, so that the total number of trainable layers matches the MLP.
We use $\sigma_F=1.0$, $n=W=80$, and $H=5$.

\paragraph{QRes (quadratic residual network).}
QRes~\cite{Bu2021QRes} replaces each linear layer with a quadratic residual
unit. For an input $x$ to a layer with weight matrices
$W_1,W_2\in\mathbb{R}^{d_\mathrm{out}\times d_\mathrm{in}}$ (both bias-free)
and a shared bias $b\in\mathbb{R}^{d_\mathrm{out}}$, the layer outputs
\begin{equation}
    \mathrm{QResLayer}(x) \;=\; (W_1 x)\odot(W_2 x) + W_1 x + b,
\end{equation}
where $\odot$ denotes the elementwise product. We follow the official
reference implementation,\footnote{\url{https://github.com/jayroxis/qres}}
in which the bias is added once alongside the linear shortcut and the
activation is applied after the layer rather than inside it. The full
backbone stacks $H=5$ such QRes layers of width $W=80$, each followed by a
$\tanh$ activation, with a final linear head. The two weight matrices per
layer give QRes roughly twice the parameter count of the MLP at the same
width and depth.

\paragraph{PirateNet.}
PirateNet~\cite{Wang2024PirateNet} combines a fixed random Fourier embedding
with adaptive residual blocks and a learnable identity-initialized skip
parameter. We mirror the official jaxpi reference
implementation.\footnote{\url{https://github.com/PredictiveIntelligenceLab/jaxpi/tree/pirate}}
Let $B\in\mathbb{R}^{d_\mathrm{in}\times (d_e/2)}$ be a fixed Gaussian kernel
with entries i.i.d.\ $\mathcal{N}(0,s^{2})$. The Fourier embedding is
\begin{equation}
    \gamma(x) \;=\; \big[\cos(B x);\ \sin(B x)\big] \;\in\;
    \mathbb{R}^{d_e}.
\end{equation}
Two parallel projections produce the gating tensors
$U=\tanh(W_U\,\gamma(x))$ and $V=\tanh(W_V\,\gamma(x))\in\mathbb{R}^{W}$.
Each of the $L$ residual blocks operates at the embedding dimension $d_e$
and, for input $x_\ell$, computes
\begin{align}
    f_1 &= \tanh(W_1^{(\ell)}\, x_\ell), &
    z_1 &= f_1\odot U + (1-f_1)\odot V, \\
    f_2 &= \tanh(W_2^{(\ell)}\, z_1), &
    z_2 &= f_2\odot U + (1-f_2)\odot V, \\
    h_\ell &= \tanh(W_3^{(\ell)}\, z_2), &
    x_{\ell+1} &= \alpha_\ell\, h_\ell + (1-\alpha_\ell)\, x_\ell,
\end{align}
where $\alpha_\ell\in\mathbb{R}$ is a learnable scalar initialized to zero so
that each block starts as the identity, the property that enables stable
deep training. A final linear head is applied to $x_{L}$. We use $L=H=5$
blocks, bottleneck width $W=80$, embedding dimension $d_e=80$ (so the kernel
$B$ has shape $d_\mathrm{in}\times 40$), and embedding scale $s=1.0$. The
additional embedding, two gating projections, and three weight matrices per
block give PirateNet a parameter count larger than the MLP.

\paragraph{KAN (Kolmogorov--Arnold network).}
KAN~\cite{Liu2024KAN} replaces every connection of the MLP with a learnable
univariate function. We use the efficient parameterization
of~\citet{Liu2024KAN},\footnote{\url{https://github.com/Blealtan/efficient-kan}}
in which each scalar connection $\phi_{ij}$ is parameterized as a sum of a
SiLU residual term and a B-spline expansion,
\begin{equation}
    \phi_{ij}(z) \;=\;
    w^{\mathrm{base}}_{ij}\,\mathrm{SiLU}(z) +
    w^{\mathrm{spline}}_{ij} \sum_{k=1}^{G+p} c^{(ij)}_{k}\, B_k(z),
\end{equation}
where $\{B_k\}$ are B-spline basis functions of order $p$ over a fixed
one-dimensional grid of size $G$, and
$w^{\mathrm{base}}_{ij}$, $w^{\mathrm{spline}}_{ij}$, and the spline
coefficients $c^{(ij)}_{k}$ are trainable. We use $G=5$, $p=3$, and stack
$H=5$ KAN layers of width $W=80$. The spline grid for the first layer
(which sees raw problem coordinates) is set per-input-dimension to cover the
training-domain bounds of the corresponding benchmark; subsequent layers,
whose inputs are bounded activations, use the default range $[-1,1]$.
Because each connection stores $G+p+1$ trainable parameters, the KAN backbone
has roughly an order of magnitude more parameters than the MLP at the same
width and depth.

\begin{sc}
\begin{table}[h]
\centering
\small
\caption{Trainable parameter counts of the alternative backbones at the
configuration used in Appendix~\ref{appendix:backbone_ablation}
($H=5$, $W=80$; $d_\mathrm{in}=3$, $d_\mathrm{out}=2$ corresponding to the
Burgers and $\lambda$--$\omega$ RD benchmarks).
The Allen--Cahn benchmark has $d_\mathrm{out}=1$, which differs only in the
final-layer parameter count and changes each entry by less than $0.1\%$.}
\label{tab:backbone_params}
\setlength{\tabcolsep}{12pt}
\begin{tabular}{lr}
\toprule
Backbone & \#Trainable parameters \\
\midrule
MLP        & 26{,}402  \\
FLS    & 26{,}402  \\
QRes       & 52{,}242  \\
PirateNet  & 110{,}327 \\
KAN        & 260{,}000 \\
\bottomrule
\end{tabular}
\end{table}
\end{sc}

\subsection{B-PINN}
B-PINN (Bayesian Physics-Informed Neural Network) is implemented as a baseline in this work. B-PINN replaces each deterministic linear layer in the shared MLP with a Bayesian linear layer whose weights and biases follow a factorized Gaussian variational posterior. During training, weights are sampled via the reparameterization trick, and the objective augments the data and PDE residual losses with a KL divergence to a zero-mean Gaussian prior. In our implementation, the prior is $N(0,1)$ and the KL term is weighted by $\lambda_{KL} = 10^{-6}$. At evaluation time, we use the posterior mean (i.e., the learned $\mu$ parameters) to produce deterministic predictions for reporting rMAE and rMSE.

\subsection{Loss functions}
\paragraph{Mean squared error.} The standard PINN and the proposed naPINN employ the mean squared error (MSE) loss, with respect to a residual $r_i$ on a measurement point.
\begin{equation}
    \ell_\mathrm{MSE}(r_i) = r_i^2.
\end{equation}
This loss corresponds to the negative log-likelihood of a Gaussian noise model and strongly penalizes large residuals.

\paragraph{Mean absolute error.} LAD-PINN replaces the quadratic data loss with an $\ell_1$ loss to improve robustness against large residuals:
\begin{equation}
    \ell_\mathrm{L1}(r_i) = |r_i|.
\end{equation}
Compared to MSE, the linear growth of the $\ell_1$ loss reduces sensitivity to large residuals but implicitly assumes a Laplace noise model with a fixed scale parameter.

\paragraph{$q$-Gaussian log-likelihood.} OrPINN adopts a Tsallis $q$-Gaussian likelihood as a flexible heavy-tailed noise model. The probability density function is defined as
\begin{equation}
    p_q(r) \propto [1+(q-1) \beta r^2]^{-\frac{1}{q-1}},
\end{equation}
which yields the per-point negative log-likelihood:
\begin{equation}
    \ell_q(r_i) = \frac{1}{q-1} \log (1+(q-1) \beta r_i^2).
\end{equation}
Following OrPINN, the scale parameter is fixed as
\begin{equation}
    \beta = \frac{1}{2(3-q)},
\end{equation}
which normalizes the variance of the distribution. In the limit $q \rightarrow 1$, the $q$-Gaussian loss smoothly recovers a quadratic loss. In our experiments, we evaluate OrPINN with $q=1.9$ and $q=2.9$.

\subsection{Evaluation metrics}
Model performance is evaluated using the relative mean absolute error (rMAE) and relative root mean squared error (rMSE), defined as
\begin{align}
\mathrm{rMAE} &= \frac{\|\hat{\mathbf{u}} - \mathbf{u}\|_1}{\|\mathbf{u}\|_1}, \\
\mathrm{rMSE} &= \frac{\|\hat{\mathbf{u}} - \mathbf{u}\|_2}{\|\mathbf{u}\|_2},
\end{align}
where $\hat{\mathbf{u}}$ and $\mathbf{u}$ denote the predicted and ground-truth solution fields evaluated on a dense reference grid. These normalized metrics enable fair comparison across different PDE systems and noise levels.

\section{Broader impacts}

This work is intended to improve the robustness of physics-informed learning in inverse PDE problems where measurements may be noisy, corrupted, or affected by sensor faults.
Potential positive impacts include more reliable reconstruction of physical states from imperfect observations and improved diagnosis of unreliable measurements in scientific and engineering workflows.
Such capabilities may be useful in settings such as environmental monitoring, industrial sensing, and simulation-assisted system identification.

The method also has limitations and possible negative consequences if used without adequate validation.
In safety-critical deployments, an incorrectly calibrated reliability gate could downweight rare but valid physical events, or could give users unwarranted confidence in predictions made from corrupted measurements.
Since our experiments are controlled simulated stress tests rather than deployment-scale real-world studies, naPINN should not be used as a stand-alone decision system in high-stakes applications without domain-specific validation, uncertainty assessment, and expert oversight.
We do not use human-subject data, personal data, scraped data, or pretrained generative models, and we do not identify a direct path to harms such as surveillance, deception, or discrimination.

\section{Licenses and assets}

This paper does not use existing public datasets, scraped data, pretrained models, or human-subject data.
The benchmark observations are synthetically generated from the PDE systems and numerical procedures described in Appendix~\ref{appendix:DatasetDetails}.
Prior methods and benchmark formulations are credited through citations in the main paper and appendix.

If code or generated benchmark assets are released with the supplementary material, they will be provided in anonymized form for review and should include a license file, dependency information, and instructions for reproducing the reported experiments.
The released code should specify its software license, and any released generated data should specify its terms of use.
No third-party asset is redistributed beyond standard open-source software dependencies required to run the experiments.


\newpage

\end{document}